\documentclass{article}
\usepackage{amsmath,amsfonts,amssymb,amsthm}
\usepackage{graphicx}
\usepackage{subcaption}
\usepackage[margin=1.2in]{geometry}
\usepackage{xcolor}
\usepackage{url}
\usepackage{booktabs}
\usepackage[normalem]{ulem}
\usepackage{hyperref}

\usepackage{tikz}
\usetikzlibrary{decorations.pathreplacing}
\usetikzlibrary{positioning}
\usetikzlibrary{fit,shapes.geometric}
\usetikzlibrary{automata}

\newcommand{\real}{\mathbb{R}}
\newcommand{\tran}{\mathsf{T}}

\newcommand{\xk}{X_k} % we may have to change this as notation evolves
\newcommand{\bsx}{\boldsymbol{x}}
\newcommand{\bszero}{\boldsymbol{0}}
\newcommand{\bsone}{\boldsymbol{1}}

\title{Probing neural networks with t-SNE, class-specific projections
and a guided tour}
\author{Christopher R. Hoyt\\Stanford University
\and
Art B. Owen\\Stanford University}
\date{July 2021}
\begin{document}
\maketitle

\begin{abstract}
We use graphical methods to probe neural nets
that classify images.   Plots of t-SNE outputs
at successive layers in a network reveal increasingly
organized arrangement of the data points.
They can also reveal how a network can diminish or even forget
about within-class structure as the data proceeds
through layers.  We use class-specific analogues
of principal components to visualize how succeeding
layers separate the classes. These allow us to
sort images from a given class from most typical to
least typical (in the data) and they also serve as very useful projection
coordinates for data visualization.  We find them especially
useful when defining versions guided tours for
animated data visualization.
\end{abstract}

\section{Introduction}

Probing is a term used to describe certain
exploratory analyses of neural networks.
It allows the user to see what happens as the
number of layers used \cite{alai:beng:2016}
or amount of training time taken
\cite{papy:han:dono:2020} increases.
Nguyen et al.~\cite{nguy:ragh:korn:2020}
consider changes to both depth and breadth
of networks.
One could also study changes to a network
as the volume of training data increases.

Alain and Bengio \cite{alai:beng:2016} investigate networks
of $L$ layers by passing the output of
the first $\ell< L$ layers through a specially retrained
softmax layer as the $\ell+1$'st and final layer.
They can then study the progress that the network
makes as layers are added.  They have the interesting
observation that the information content in layer $\ell+1$
can be less than what is in layer $\ell$ but not more,
and yet accuracy may improve because the information has a better
representation at later layers.

In this work, we present some tools for
graphically probing neural nets.
The nets we study as examples are designed to classify
an image into one of $K$ classes.
Instead of passing the output of an intermediate
layer through a special softmax layer
we explore each layer's output visually.
First, we run t-SNE \cite{vand:hint:2008} on the
outputs of  intermediate layers. Sometimes we
see interesting phenomena in the intermediate
layers that have disappeared by the final layers.
For instance, within-class clustering could
be evident in the early layers but forgotten
by the final layer without harming the final
classification accuracy that the
network was trained to optimize.

Next, we look at principal components (PC) projections of
intermediate layer results. By watching how those change
from layer to layer we can see how the group separations
evolve.  The PC projections have the advantage that we
can project additional points, such as held out data points,
onto their coordinates.  This is different from t-SNE where
including additional points would change the mapping of all the points.
Also, PC mappings are deterministic given the data, and hence more reproducible,
while the t-SNE images are randomly generated.
We do not always see the neural collapse phenomenon
from \cite{papy:han:dono:2020}.  That analysis predicts
that the neuron outputs for different categories will
end up as roughly spherical clusters at the corners
of a simplex.  Instead of clusters we often see ellipsoids
extending from such a corner towards the origin.
Neural collapse predicts an angle of about $90$ degrees
between clusters. We see that this emerges as layers progress, but in early layers than angle can be much less.
% (modified this sentence a bit).
 In intermediate
layers we often see angles much greater than $90$ degrees
as if hard to separate classes are being treated as opposites.

When there are many classes, principal components trained
on all of them can make it harder to spot patterns among a few
commonly confused classes.  We could get better resolution by
defining principal components tuned to one subset
of classes. That, however, requires a separate PCA analysis for
every subset of variables.  We have found it beneficial to define
one class-specific vector per output class in a way
that is analagous to the PC eigenvectors.
Then each subset of categories comes with a prescribed
set of vectors for projection.
Similar class-specfic vectors were previously studied by Krzanowski \cite{krza:1979} but
that work was not about visualization.
We can also use these vectors to rank input images from most
typical to least typical members of their respective classes.
Note that typical is defined with respect to the set of images
gathered for training and those images might not remain typical when an
algorithm is deployed.

We also consider some animated tours
\cite{asim:1985, buja:cook:asim:hurl:2005}
%\textcolor{blue}{Ok.  I thought from your prior description that the
%web doc was where you got the technical details.  If that is true, please
%revert to citing it.  For now, I'll take it out.} \textcolor{red}{\cite{buja:cook:asim:hurl:2005} is the reference that I used for the computations themselves. The other paper \cite{buja:etal:2004} also makes references to tours, but doesn't cover the gritty details.}
of data at each level where we probe.
Li et al.~\cite{li:zhao:sche:2020} previously used grand tours on the
output of a neural network.
In the terminology of \cite{buja:cook:asim:hurl:2005} our tours are
guided tours: we select pairs of interesting projections and
interpolate between them.
We develop guided tours based on class-specific projections
of data including looks at intermediate layers.

We use two well studied neural networks as running examples.
One is the CIFAR-10 dataset \cite{kriz:2009}
of sixty thousand 32 by 32 color images selected from 10 classes consisting of four
mechanical groups (e.g., airplanes and automobiles) and six animal groups (e.g.,
cats and dogs).
For this we use the network and trained weights from
https://github.com/geifmany/cifar-vgg which utilizes the VGG model with
15 layers.  VGG stands for the `visual geometry group' at Oxford who developed the architecture.
The other example is the fashion-MNIST data with sixty thousand 28 by 28 grayscale images selected from 10 classes of clothing items (like trousers and pullovers), which was then subsequently padded out with zeros along the edges to become 32 by 32 images with three channels.
For this we consider the ResNet50 network architecture \cite{he:zhan:ren:sun:2015}
with weights initialized from ImageNet pretraining \cite{deng:dong:soch:li:li:fei:2009}.

The outline of this paper is as follows.
Section~\ref{sec:back} gives background
about the neural networks we explore.
Section~\ref{sec:tsne} shows a use of t-SNE
for probing.  We can see how class
separation progresses as the layers increase.
Section~\ref{sec:classv} defines our class-specific vectors and
compares plots in those
coordinates to principal components plots.
Section \ref{sec:classt} describes our class-specific
tours.  It include still shots of the tours and
a link to animated versions. Section \ref{sec:ingroup} describes how our class-specific vectors can potentially identify separations of groups within a single classification.

\section{Image data sets, architectures, and }\label{sec:back}

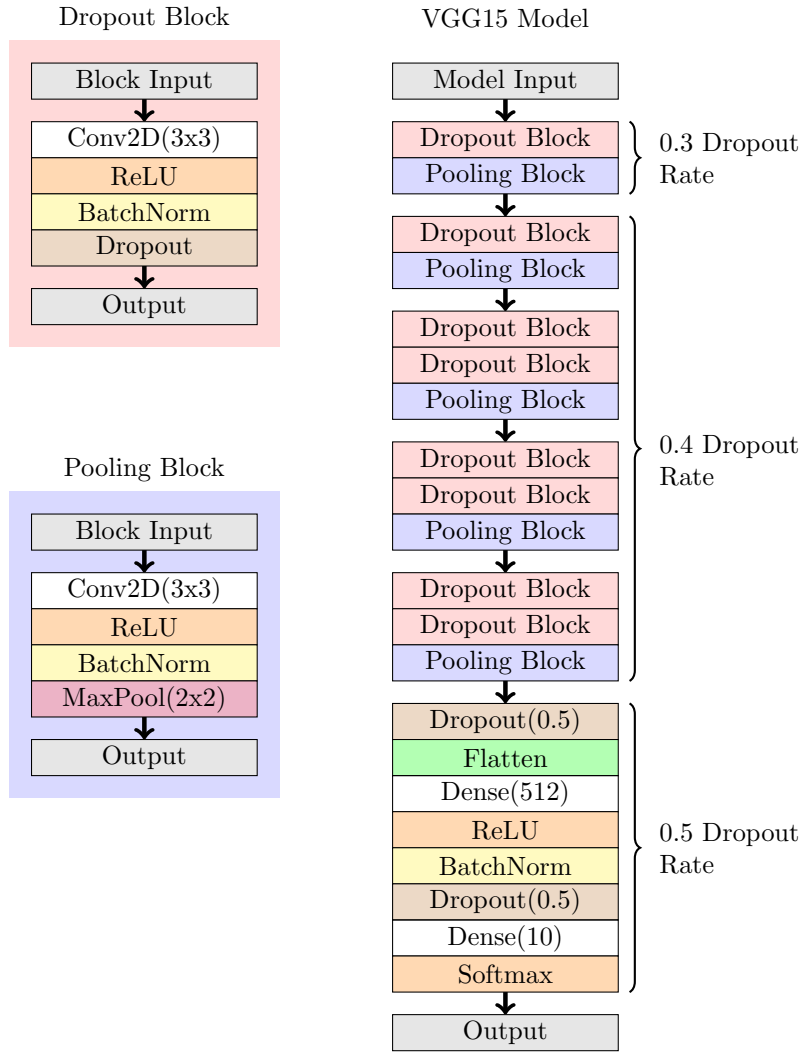
\begin{figure}
\begin{subfigure}{.9\textwidth}
\centering
\begin{tikzpicture}[scale=0.6, rotate=-90]
%\draw ( 0, 0) rectangle ( 0.9, 4) node[pos=.5, rotate=90] { Conv2D };

\def\w{5}

\def\h{0.8}
\def\blockspacing{0}
\def\groupspacing{0.5}

\fill[red!15] (-\h-\groupspacing-0.5,-0.5) rectangle ( 5*\h+\groupspacing+0.5, \w+0.5);
\node at (-\h-\groupspacing-1, \w/2) { Dropout Block };

\draw[fill=  gray!20] ( -\h-\groupspacing , 0) rectangle ( -\groupspacing-\blockspacing, \w) node[pos=.5] { Block Input };

\draw [->, line width=0.6mm] (-\groupspacing-\blockspacing, \w/2) -- (0,\w/2);

\draw[fill=    white] (   0  , 0) rectangle (  \h-\blockspacing, \w) node[pos=.5] { Conv2D(3x3) };
\draw[fill=orange!30] (  \h  , 0) rectangle (2*\h-\blockspacing, \w) node[pos=.5] { ReLU        };
\draw[fill=yellow!30] (2*\h  , 0) rectangle (3*\h-\blockspacing, \w) node[pos=.5] { BatchNorm   };
\draw[fill= brown!30] (3*\h  , 0) rectangle (4*\h-\blockspacing, \w) node[pos=.5] { Dropout     };

\draw [->, line width=0.6mm] (4*\h-\blockspacing, \w/2) -- (4*\h+\groupspacing,\w/2);

\draw[fill=  gray!20] (4*\h+\groupspacing, 0) rectangle ( 5*\h+\groupspacing, \w) node[pos=.5] { Output };

%%%%%%%%%%%%%%%%%%%%%%%%%%%%%%%%%%%%%%%%%%%%%%%%%%%%%%%%%%%%%%%%%%%%%%%%%%%%%%%%%%%%%%

\def\yPoolingVGG{10}

\fill[blue!15] (\yPoolingVGG-\h-\groupspacing-0.5,-0.5) rectangle (\yPoolingVGG+5*\h+\groupspacing+0.5, \w+0.5);
\node at (\yPoolingVGG-\h-\groupspacing-1, \w/2) { Pooling Block };

\draw[fill=  gray!20] (\yPoolingVGG-\h-\groupspacing, 0) rectangle (\yPoolingVGG-\groupspacing-\blockspacing, \w) node[pos=.5] { Block Input };

\draw [->, line width=0.6mm] (\yPoolingVGG-\groupspacing-\blockspacing, \w/2) -- (\yPoolingVGG,\w/2);

\draw[fill=    white] (\yPoolingVGG+    0  , 0) rectangle (\yPoolingVGG+ 1*\h-\blockspacing, \w) node[pos=.5] { Conv2D(3x3)  };
\draw[fill=orange!30] (\yPoolingVGG+   \h  , 0) rectangle (\yPoolingVGG+ 2*\h-\blockspacing, \w) node[pos=.5] { ReLU         };
\draw[fill=yellow!30] (\yPoolingVGG+ 2*\h  , 0) rectangle (\yPoolingVGG+ 3*\h-\blockspacing, \w) node[pos=.5] { BatchNorm    };
\draw[fill=purple!30] (\yPoolingVGG+ 3*\h  , 0) rectangle (\yPoolingVGG+ 4*\h-\blockspacing, \w) node[pos=.5] { MaxPool(2x2) };

\draw [->, line width=0.6mm] (\yPoolingVGG+4*\h-\blockspacing, \w/2) -- (\yPoolingVGG+4*\h+\groupspacing, \w/2);

\draw[fill=  gray!20] (\yPoolingVGG+4*\h+\groupspacing, 0) rectangle (\yPoolingVGG+5*\h+\groupspacing, \w) node[pos=.5] { Output };

%%%%%%%%%%%%%%%%%%%%%%%%%%%%%%%%%%%%%%%%%%%%%%%%%%%%%%%%%%%%%%%%%%%%%%%%%%%%%%%%%%%%%%%%%%%%
\def\xMainVGG{8}

\node at (-\h-\groupspacing-1,\xMainVGG+\w/2) {VGG15 Model};

\draw[fill=  gray!20] ( -\h-\groupspacing  , \xMainVGG) rectangle (-\groupspacing-\blockspacing, \xMainVGG+\w) node[pos=.5] { Model Input };

\draw [->, line width=0.6mm] (-\groupspacing-\blockspacing, \xMainVGG+\w/2) -- (0,\xMainVGG+\w/2);

\draw[fill=   red!15] ( 0  , \xMainVGG) rectangle ( \h-\blockspacing, \xMainVGG+\w) node[pos=.5] { Dropout Block };
\draw[fill=  blue!15] ( \h  , \xMainVGG) rectangle ( 2*\h-\blockspacing, \xMainVGG+\w) node[pos=.5] { Pooling Block };

\draw [->, line width=0.6mm] (2*\h-\blockspacing, \xMainVGG+\w/2) -- (2*\h+\groupspacing,\xMainVGG+\w/2);

\draw[fill=   red!15] (2*\h+\groupspacing, \xMainVGG) rectangle (3*\h+\groupspacing-\blockspacing, \xMainVGG+\w) node[pos=.5] { Dropout Block };
\draw[fill=  blue!15] (3*\h+\groupspacing, \xMainVGG) rectangle (4*\h+\groupspacing-\blockspacing, \xMainVGG+\w) node[pos=.5] { Pooling Block };

\draw [->, line width=0.6mm] (4*\h+\groupspacing-\blockspacing, \xMainVGG+\w/2) -- (4*\h+2*\groupspacing,\xMainVGG+\w/2);

\draw[fill=   red!15] (4*\h+2*\groupspacing, \xMainVGG) rectangle (5*\h+2*\groupspacing-\blockspacing, \xMainVGG+\w) node[pos=.5] { Dropout Block };
\draw[fill=   red!15] (5*\h+2*\groupspacing, \xMainVGG) rectangle (6*\h+2*\groupspacing-\blockspacing, \xMainVGG+\w) node[pos=.5] { Dropout Block };
\draw[fill=  blue!15] (6*\h+2*\groupspacing, \xMainVGG) rectangle (7*\h+2*\groupspacing-\blockspacing, \xMainVGG+\w) node[pos=.5] { Pooling Block };

\draw [->, line width=0.6mm] (7*\h+2*\groupspacing-\blockspacing, \xMainVGG+\w/2) -- (7*\h+3*\groupspacing,\xMainVGG+\w/2);

\draw[fill=   red!15] (7*\h+3*\groupspacing, \xMainVGG) rectangle (8*\h+3*\groupspacing-\blockspacing, \xMainVGG+\w) node[pos=.5] { Dropout Block };
\draw[fill=   red!15] (8*\h+3*\groupspacing, \xMainVGG) rectangle (9*\h+3*\groupspacing-\blockspacing, \xMainVGG+\w) node[pos=.5] { Dropout Block };
\draw[fill=  blue!15] (9*\h+3*\groupspacing, \xMainVGG) rectangle (10*\h+3*\groupspacing-\blockspacing, \xMainVGG+\w) node[pos=.5] { Pooling Block };

\draw [->, line width=0.6mm] (10*\h+3*\groupspacing-\blockspacing, \xMainVGG+\w/2) -- (10*\h+4*\groupspacing,\xMainVGG+\w/2);

\draw[fill=   red!15] (10*\h+4*\groupspacing, \xMainVGG) rectangle (11*\h+4*\groupspacing-\blockspacing, \xMainVGG+\w) node[pos=.5] { Dropout Block };
\draw[fill=   red!15] (11*\h+4*\groupspacing, \xMainVGG) rectangle (12*\h+4*\groupspacing-\blockspacing, \xMainVGG+\w) node[pos=.5] { Dropout Block };
\draw[fill=  blue!15] (12*\h+4*\groupspacing, \xMainVGG) rectangle (13*\h+4*\groupspacing-\blockspacing, \xMainVGG+\w) node[pos=.5] { Pooling Block };

\draw [->, line width=0.6mm] (13*\h+4*\groupspacing-\blockspacing, \xMainVGG+\w/2) -- (13*\h+5*\groupspacing,\xMainVGG+\w/2);

\draw[fill= brown!30] (13*\h+5*\groupspacing, \xMainVGG) rectangle (14*\h+5*\groupspacing-\blockspacing, \xMainVGG+\w) node[pos=.5] { Dropout(0.5)        };
\draw[fill= green!30] (14*\h+5*\groupspacing, \xMainVGG) rectangle (15*\h+5*\groupspacing-\blockspacing, \xMainVGG+\w) node[pos=.5] { Flatten             };
\draw                 (15*\h+5*\groupspacing, \xMainVGG) rectangle (16*\h+5*\groupspacing-\blockspacing, \xMainVGG+\w) node[pos=.5] { Dense(512)          };
\draw[fill=orange!30] (16*\h+5*\groupspacing, \xMainVGG) rectangle (17*\h+5*\groupspacing-\blockspacing, \xMainVGG+\w) node[pos=.5] { ReLU                };
\draw[fill=yellow!30] (17*\h+5*\groupspacing, \xMainVGG) rectangle (18*\h+5*\groupspacing-\blockspacing, \xMainVGG+\w) node[pos=.5] { BatchNorm           };
\draw[fill= brown!30] (18*\h+5*\groupspacing, \xMainVGG) rectangle (19*\h+5*\groupspacing-\blockspacing, \xMainVGG+\w) node[pos=.5] { Dropout(0.5)        };
\draw                 (19*\h+5*\groupspacing, \xMainVGG) rectangle (20*\h+5*\groupspacing-\blockspacing, \xMainVGG+\w) node[pos=.5] { Dense(10)           };
\draw[fill=orange!30] (20*\h+5*\groupspacing, \xMainVGG) rectangle (21*\h+5*\groupspacing-\blockspacing, \xMainVGG+\w) node[pos=.5] { Softmax             };

\draw [->, line width=0.6mm] (21*\h+5*\groupspacing-\blockspacing, \xMainVGG+\w/2) -- (21*\h+6*\groupspacing,\xMainVGG+\w/2);

\draw[fill=  gray!20] (21*\h+6*\groupspacing, \xMainVGG) rectangle (22*\h+6*\groupspacing, \xMainVGG+\w) node[pos=.5] { Output };

\draw [thick,decorate,decoration={brace,amplitude=4pt},yshift=0.25cm] (0,\xMainVGG+\w) -- (2*\h-\blockspacing,\xMainVGG+\w);
\node[text width=2.5cm] at (\h-\blockspacing/2, \xMainVGG+\w+3) {$0.3$ Dropout Rate};

\draw [thick,decorate,decoration={brace,amplitude=4pt},yshift=0.25cm] (2*\h+\groupspacing,\xMainVGG+\w) -- (13*\h+4*\groupspacing-\blockspacing,\xMainVGG+\w);
\node[text width=2.5cm] at (7.5*\h+2.5*\groupspacing+0.5*\groupspacing-0.5*\blockspacing, \xMainVGG+\w+3) {$0.4$ Dropout Rate};

\draw [thick,decorate,decoration={brace,amplitude=4pt},yshift=0.25cm] (13*\h+5*\groupspacing,\xMainVGG+\w) -- (21*\h+5*\groupspacing-\blockspacing,\xMainVGG+\w);
\node[text width=2.5cm] at (17*\h+5*\groupspacing-0.5*\blockspacing, \xMainVGG+\w+3) {$0.5$ Dropout Rate};
\end{tikzpicture}
\end{subfigure}
\caption{\label{fig:VGG15Model} The model architecture for the VGG15 model. The model consists of thirteen convolution layers and two dense layers, creating fifteen natural checkpoints for the model overall. Refer to https://github.com/geifmany/cifar-vgg for more specific details on the model architecture.
}
\end{figure}

\begin{figure}
\begin{subfigure}{.475\textwidth}
\centering
\hspace*{-1cm}
\includegraphics[width=1.1\hsize]{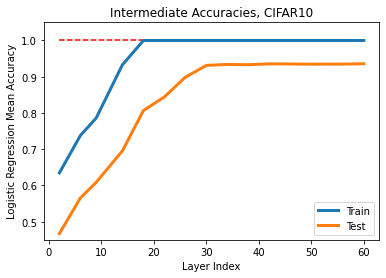}
\end{subfigure}
\begin{subfigure}{.475\textwidth}
\centering
\includegraphics[width=1.1\hsize]{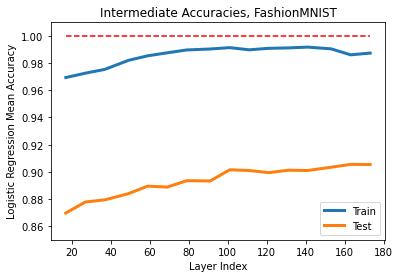}
\end{subfigure}
\caption{\label{fig:intermediateAccuracy}
Intermediate accuracies of the models used for the CIFAR10 and FashionMNIST datasets using a logistic regression trained on a random subset of $n=10,000$ samples from the training dataset. We plot the accuracy of the model versus layer
with one curve for the training inputs and another for the test inputs.
%(Side note: the earlier plots looked at training on the test dataset, and then evaluate on the test data, which is why these plots look a bit different from before)
}
\end{figure}

The network we use for CIFAR-10
from \url{https://github.com/geifmany/cifar-vgg} has
thirteen convolution layers and two fully connected layers
that we study.
Those are layers
$$\ell = 2, 6, 9, 14, 18, 22, 26, 30, 34, 38, 42, 46, 50, 56, 60$$ of that network.
Figure \ref{fig:VGG15Model} shows their positions in the overall architecture.
We probed the network outputs at these layers by passing the output from
layer $\ell$ through a specially trained final softmax layer.
Figure \ref{fig:intermediateAccuracy} plots the accuracy
attained this way versus the layer indices.
What we see is that the accuracy on test data
increased up to layer 30,  and subsequent layers did not improve it by  much.
The accuracy on the training data saturated at $100$\% by layer 18.
In a Shapley value analysis (to be reported elsewhere) we found that
of these layers, layer $2$ was most important for ultimate accuracy.
At layer 2, the network is not anywhere near it's final accuracy.

\begin{figure}
\begin{subfigure}{.9\textwidth}
\centering
\begin{tikzpicture}[scale=0.6, rotate=-90]

\def\w{5}
\def\h{0.8}
\def\blockspacing{0}
\def\groupspacing{0.5}
\def\grouphorispacing{0.5}

\def\xConvResnet{0}
\def\yConvResnet{15}

\node at (\yConvResnet+-1-\h-\groupspacing-\blockspacing,\xConvResnet+ \w + \grouphorispacing/2) { Convolution Block };
\fill[red!15] (\yConvResnet+-0.5-\h-\groupspacing-\blockspacing,\xConvResnet+-0.5) rectangle (\yConvResnet+ 0.5+11*\h+4*\groupspacing-\blockspacing,\xConvResnet+ 0.5+2*\w+\grouphorispacing);

\draw[fill=gray!30] (\yConvResnet+ -\h-\groupspacing-\blockspacing,\xConvResnet+ 0) rectangle (\yConvResnet+ -\groupspacing-\blockspacing,\xConvResnet+ 2*\w+\grouphorispacing) node[pos=0.5] { Block Input };

\draw [->, line width=0.6mm] (\yConvResnet+-\groupspacing-\blockspacing,\xConvResnet+ \w/2) -- (\yConvResnet+0,\xConvResnet+\w/2);
\draw [->, line width=0.6mm] (\yConvResnet+-\groupspacing-\blockspacing,\xConvResnet+ 3*\w/2 + \grouphorispacing) -- (\yConvResnet+0,\xConvResnet+3*\w/2 + \grouphorispacing);

\draw[fill=white]     (\yConvResnet+ 0,\xConvResnet+ 0) rectangle (\yConvResnet+ \h-\blockspacing,\xConvResnet+ \w) node[pos=.5] { Conv2D(1x1) };
\draw[fill=yellow!30] (\yConvResnet+ \h,\xConvResnet+ 0) rectangle (\yConvResnet+ 2*\h-\blockspacing,\xConvResnet+ \w) node[pos=.5] { BatchNorm };
\draw[fill=orange!30] (\yConvResnet+ 2*\h,\xConvResnet+ 0) rectangle (\yConvResnet+ 3*\h-\blockspacing,\xConvResnet+ \w) node[pos=.5] { ReLU };

\draw [->, line width=0.6mm] (\yConvResnet+3*\h-\blockspacing,\xConvResnet+ \w/2) -- (\yConvResnet+3*\h+\groupspacing,\xConvResnet+\w/2);

\draw[fill=white]     (\yConvResnet+3*\h+\groupspacing,\xConvResnet+ 0) rectangle (\yConvResnet+4*\h+\groupspacing-\blockspacing,\xConvResnet+ \w) node[pos=.5] { Conv2D(3x3) };
\draw[fill=yellow!30] (\yConvResnet+4*\h+\groupspacing,\xConvResnet+ 0) rectangle (\yConvResnet+5*\h+\groupspacing-\blockspacing,\xConvResnet+ \w) node[pos=.5] { BatchNorm };
\draw[fill=orange!30] (\yConvResnet+5*\h+\groupspacing,\xConvResnet+ 0) rectangle (\yConvResnet+6*\h+\groupspacing-\blockspacing,\xConvResnet+ \w) node[pos=.5] { ReLU };

\draw [->, line width=0.6mm] (\yConvResnet+6*\h+\groupspacing-\blockspacing,\xConvResnet+ \w/2) -- (\yConvResnet+6*\h+2*\groupspacing,\xConvResnet+\w/2);

\draw[fill=white] (\yConvResnet+6*\h+2*\groupspacing,\xConvResnet+ 0) rectangle (\yConvResnet+7*\h+2*\groupspacing-\blockspacing,\xConvResnet+ \w) node[pos=.5] { Conv2D(1x1) };
\draw[fill=yellow!30] (\yConvResnet+7*\h+2*\groupspacing,\xConvResnet+ 0) rectangle (\yConvResnet+8*\h+2*\groupspacing-\blockspacing,\xConvResnet+ \w) node[pos=.5] { BatchNorm };

\draw [->, line width=0.6mm] (\yConvResnet+8*\h+2*\groupspacing-\blockspacing,\xConvResnet+ \w/2) -- (\yConvResnet+8*\h+3*\groupspacing,\xConvResnet+\w/2);

%\draw [->, line width=0.6mm] (\yConvResnet+2.9, 2.5) -- (\yConvResnet+3.5,\xConvResnet+2.5);
\draw[fill=white] (\yConvResnet+ 0,\xConvResnet+ \w+\grouphorispacing) rectangle (\yConvResnet+\h-\blockspacing,\xConvResnet+ 2*\w+\grouphorispacing) node[pos=.5] { Conv2D(1x1) };
\draw[fill=yellow!30] (\yConvResnet+\h,\xConvResnet+ \w+\grouphorispacing) rectangle (\yConvResnet+2*\h-\blockspacing,\xConvResnet+ 2*\w+\grouphorispacing) node[pos=.5] { BatchNorm };

\draw [->, line width=0.6mm] (\yConvResnet+2*\h-\blockspacing,\xConvResnet+ 1.5*\w+\grouphorispacing) -- (\yConvResnet+8*\h+3*\groupspacing,\xConvResnet+1.5*\w+\grouphorispacing);

\draw[fill=green!30] (\yConvResnet+8*\h+3*\groupspacing,\xConvResnet+ 0) rectangle (\yConvResnet+9*\h+3*\groupspacing-\blockspacing,\xConvResnet+ 2*\w+\grouphorispacing) node[pos=.5] { Add };
\draw[fill=orange!30] (\yConvResnet+9*\h+3*\groupspacing,\xConvResnet+ 0) rectangle (\yConvResnet+10*\h+3*\groupspacing-\blockspacing,\xConvResnet+ 2*\w+\grouphorispacing) node[pos=.5] { ReLU };

\draw [->, line width=0.6mm] (\yConvResnet+10*\h+3*\groupspacing-\blockspacing,\xConvResnet+ \w+\grouphorispacing/2) -- (\yConvResnet+10*\h+4*\groupspacing,\xConvResnet+\w+\grouphorispacing/2);

\draw[fill=gray!30] (\yConvResnet+10*\h+4*\groupspacing,\xConvResnet+ 0) rectangle (\yConvResnet+11*\h+4*\groupspacing-\blockspacing,\xConvResnet+ 2*\w+\grouphorispacing) node[pos=0.5] { Block Output };

%%%%%%%%%%%%%%%%%%%%%%%%%%%%%%%%%%%%%%%%%%%%%%%%%%%%%%%%%%%%%%%%%%%%%%%%%%%%%

\node at (-1-\h-\groupspacing-\blockspacing, \w + \grouphorispacing/2) { Identity Block };
\fill[blue!15] (-0.5-\h-\groupspacing-\blockspacing,-0.5) rectangle ( 0.5+11*\h+4*\groupspacing-\blockspacing, 0.5+2*\w+\grouphorispacing);

\draw[fill=gray!30] ( -\h-\groupspacing-\blockspacing, 0) rectangle ( -\groupspacing-\blockspacing, 2*\w+\grouphorispacing) node[pos=0.5] { Block Input };

\draw [->, line width=0.6mm] (-\groupspacing-\blockspacing, \w/2) -- (0,\w/2);
\draw [->, line width=0.6mm] (-\groupspacing-\blockspacing, 3*\w/2 + \grouphorispacing) -- (0,3*\w/2 + \grouphorispacing);

\draw[fill=white]     ( 0, 0) rectangle ( \h-\blockspacing, \w) node[pos=.5] { Conv2D(1x1) };
\draw[fill=yellow!30] ( \h, 0) rectangle ( 2*\h-\blockspacing, \w) node[pos=.5] { BatchNorm };
\draw[fill=orange!30] ( 2*\h, 0) rectangle ( 3*\h-\blockspacing, \w) node[pos=.5] { ReLU };

\draw [->, line width=0.6mm] (3*\h-\blockspacing, \w/2) -- (3*\h+\groupspacing,\w/2);

\draw[fill=white]     (3*\h+\groupspacing, 0) rectangle (4*\h+\groupspacing-\blockspacing, \w) node[pos=.5] { Conv2D(3x3) };
\draw[fill=yellow!30] (4*\h+\groupspacing, 0) rectangle (5*\h+\groupspacing-\blockspacing, \w) node[pos=.5] { BatchNorm };
\draw[fill=orange!30] (5*\h+\groupspacing, 0) rectangle (6*\h+\groupspacing-\blockspacing, \w) node[pos=.5] { ReLU };

\draw [->, line width=0.6mm] (6*\h+\groupspacing-\blockspacing, \w/2) -- (6*\h+2*\groupspacing,\w/2);

\draw[fill=white] (6*\h+2*\groupspacing, 0) rectangle (7*\h+2*\groupspacing-\blockspacing, \w) node[pos=.5] { Conv2D(1x1) };
\draw[fill=yellow!30] (7*\h+2*\groupspacing, 0) rectangle (8*\h+2*\groupspacing-\blockspacing, \w) node[pos=.5] { BatchNorm };

\draw [->, line width=0.6mm] (8*\h+2*\groupspacing-\blockspacing, \w/2) -- (8*\h+3*\groupspacing,\w/2);

%\draw [->, line width=0.6mm] (2.9, 2.5) -- (3.5,2.5);
\draw[fill=white] ( 0, \w+\grouphorispacing) rectangle (\h-\blockspacing, 2*\w+\grouphorispacing) node[pos=.5] { Conv2D(1x1) };
\draw[fill=yellow!30] (\h, \w+\grouphorispacing) rectangle (2*\h-\blockspacing, 2*\w+\grouphorispacing) node[pos=.5] { BatchNorm };

\draw [->, line width=0.6mm] (2*\h-\blockspacing, 1.5*\w+\grouphorispacing) -- (8*\h+3*\groupspacing,1.5*\w+\grouphorispacing);

\draw[fill=green!30] (8*\h+3*\groupspacing, 0) rectangle (9*\h+3*\groupspacing-\blockspacing, 2*\w+\grouphorispacing) node[pos=.5] { Add };
\draw[fill=orange!30] (9*\h+3*\groupspacing, 0) rectangle (10*\h+3*\groupspacing-\blockspacing, 2*\w+\grouphorispacing) node[pos=.5] { ReLU };

\draw [->, line width=0.6mm] (10*\h+3*\groupspacing-\blockspacing, \w+\grouphorispacing/2) -- (10*\h+4*\groupspacing,\w+\grouphorispacing/2);

\draw[fill=gray!30] (10*\h+4*\groupspacing, 0) rectangle (11*\h+4*\groupspacing-\blockspacing, 2*\w+\grouphorispacing) node[pos=0.5] { Block Output };

%%%%%%%%%%%%%%%%%%%%%%%%%%%%%%%%%%%%%%%%%%%%%%%%%%%%%%%%%%%%%%%%%%%%%%%%%%%%%

\def\xMainResnet{14}

\node at (-\h -\groupspacing - 1,\xMainResnet+\w/2) {ResNet50 Model};

\draw[fill=gray!20] (-\h -\groupspacing ,\xMainResnet+0) rectangle (-\groupspacing-\blockspacing ,\xMainResnet+\w) node[pos=0.5] { Model Input };

\draw [-> ,line width=0.6mm] (-\groupspacing-\blockspacing ,\xMainResnet+\w/2) -- (0,\xMainResnet+\w/2);

\draw                 ( 0 ,\xMainResnet+0) rectangle (\h-\blockspacing ,\xMainResnet+\w) node[pos=.5] { Conv2D(7x7) };
\draw[fill=yellow!30] ( 1*\h ,\xMainResnet+0) rectangle ( 2*\h-\blockspacing ,\xMainResnet+\w) node[pos=.5] { BatchNorm };
\draw[fill=orange!30] ( 2*\h ,\xMainResnet+0) rectangle ( 3*\h-\blockspacing ,\xMainResnet+\w) node[pos=.5] { ReLU };
\draw[fill=purple!30] ( 3*\h ,\xMainResnet+0) rectangle ( 4*\h-\blockspacing ,\xMainResnet+\w) node[pos=.5] { MaxPool(3x3) };

\draw [-> ,line width=0.6mm] (4*\h-\blockspacing ,\xMainResnet+\w/2) -- (4*\h+\groupspacing,\xMainResnet+\w/2);

\draw[fill= red!15] ( 4*\h+\groupspacing ,\xMainResnet+0) rectangle ( 5*\h+\groupspacing-\blockspacing ,\xMainResnet+\w) node[pos=.5] { Conv. Block };
\draw[fill=blue!15] ( 5*\h+\groupspacing ,\xMainResnet+0) rectangle ( 6*\h+\groupspacing-\blockspacing ,\xMainResnet+\w) node[pos=.5] { Identity Block };
\draw[fill=blue!15] ( 6*\h+\groupspacing ,\xMainResnet+0) rectangle ( 7*\h+\groupspacing-\blockspacing ,\xMainResnet+\w) node[pos=.5] { Identity Block };

\draw [-> ,line width=0.6mm] (7*\h+\groupspacing-\blockspacing ,\xMainResnet+\w/2) -- (7*\h+2*\groupspacing,\xMainResnet+\w/2);

\draw[fill= red!15] ( 7*\h+2*\groupspacing ,\xMainResnet+0) rectangle ( 8*\h+2*\groupspacing-\blockspacing ,\xMainResnet+\w) node[pos=.5] { Conv. Block };
\draw[fill=blue!15] ( 8*\h+2*\groupspacing ,\xMainResnet+0) rectangle ( 9*\h+2*\groupspacing-\blockspacing ,\xMainResnet+\w) node[pos=.5] { Identity Block };
\draw[fill=blue!15] ( 9*\h+2*\groupspacing ,\xMainResnet+0) rectangle (10*\h+2*\groupspacing-\blockspacing ,\xMainResnet+\w) node[pos=.5] { Identity Block };
\draw[fill=blue!15] (10*\h+2*\groupspacing ,\xMainResnet+0) rectangle (11*\h+2*\groupspacing-\blockspacing ,\xMainResnet+\w) node[pos=.5] { Identity Block };

\draw [-> ,line width=0.6mm] (11*\h+2*\groupspacing-\blockspacing ,\xMainResnet+\w/2) -- (11*\h+3*\groupspacing,\xMainResnet+\w/2);

\draw[fill= red!15] (11*\h+3*\groupspacing ,\xMainResnet+0) rectangle (12*\h+3*\groupspacing-\blockspacing ,\xMainResnet+\w) node[pos=.5] { Conv. Block };
\draw[fill=blue!15] (12*\h+3*\groupspacing ,\xMainResnet+0) rectangle (13*\h+3*\groupspacing-\blockspacing ,\xMainResnet+\w) node[pos=.5] { Identity Block };
\draw[fill=blue!15] (13*\h+3*\groupspacing ,\xMainResnet+0) rectangle (14*\h+3*\groupspacing-\blockspacing ,\xMainResnet+\w) node[pos=.5] { Identity Block };
\draw[fill=blue!15] (14*\h+3*\groupspacing ,\xMainResnet+0) rectangle (15*\h+3*\groupspacing-\blockspacing ,\xMainResnet+\w) node[pos=.5] { Identity Block };
\draw[fill=blue!15] (15*\h+3*\groupspacing ,\xMainResnet+0) rectangle (16*\h+3*\groupspacing-\blockspacing ,\xMainResnet+\w) node[pos=.5] { Identity Block };
\draw[fill=blue!15] (16*\h+3*\groupspacing ,\xMainResnet+0) rectangle (17*\h+3*\groupspacing-\blockspacing ,\xMainResnet+\w) node[pos=.5] { Identity Block };

\draw [-> ,line width=0.6mm] (17*\h+3*\groupspacing-\blockspacing ,\xMainResnet+\w/2) -- (17*\h+4*\groupspacing,\xMainResnet+\w/2);

\draw[fill= red!15] (17*\h+4*\groupspacing ,\xMainResnet+0) rectangle (18*\h+4*\groupspacing-\blockspacing ,\xMainResnet+\w) node[pos=.5] { Conv. Block };
\draw[fill=blue!15] (18*\h+4*\groupspacing ,\xMainResnet+0) rectangle (19*\h+4*\groupspacing-\blockspacing ,\xMainResnet+\w) node[pos=.5] { Identity Block };
\draw[fill=blue!15] (19*\h+4*\groupspacing ,\xMainResnet+0) rectangle (20*\h+4*\groupspacing-\blockspacing ,\xMainResnet+\w) node[pos=.5] { Identity Block };

\draw [-> ,line width=0.6mm] (20*\h+4*\groupspacing-\blockspacing ,\xMainResnet+\w/2) -- (20*\h+5*\groupspacing,\xMainResnet+\w/2);

\draw[fill=purple!30] (20*\h+5*\groupspacing ,\xMainResnet+0) rectangle (21*\h+5*\groupspacing-\blockspacing ,\xMainResnet+\w) node[pos=.5] { AvgPool   };
\draw                 (21*\h+5*\groupspacing ,\xMainResnet+0) rectangle (22*\h+5*\groupspacing-\blockspacing ,\xMainResnet+\w) node[pos=.5] { Dense(10) };
\draw[fill=orange!30] (22*\h+5*\groupspacing ,\xMainResnet+0) rectangle (23*\h+5*\groupspacing-\blockspacing ,\xMainResnet+\w) node[pos=.5] { Softmax   };

\draw [-> ,line width=0.6mm] (23*\h+5*\groupspacing-\blockspacing ,\xMainResnet+\w/2) -- (23*\h+6*\groupspacing,\xMainResnet+\w/2);

\draw[fill=gray!20] (23*\h+6*\groupspacing ,\xMainResnet+0) rectangle (24*\h+6*\groupspacing-\blockspacing ,\xMainResnet+\w) node[pos=0.5] { Model Output };
\end{tikzpicture}
\end{subfigure}
\caption{\label{fig:ResNetModel}
]The model architecture for the ResNet50 model. The model starts off with an initial processing section, and then proceeds with a series of 16 residual units (the identity and convolution blocks). As we progress through the blocks, we decrease the width and height of the inputs and increase the number of channels. Refer to \cite{he:zhan:ren:sun:2015} for more specific details on the model architecture.
}
\end{figure}
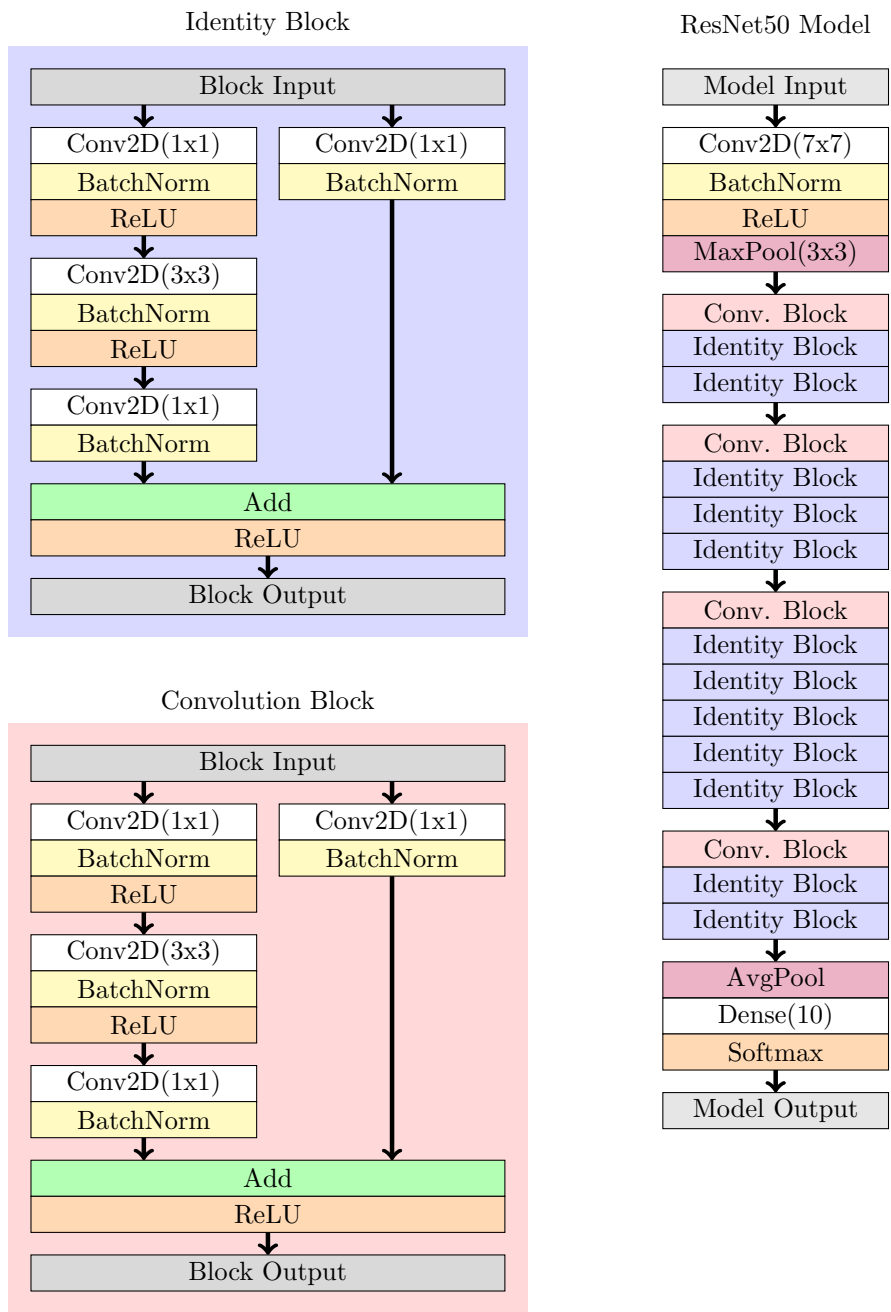

For fashion-MNIST we used the ResNet50 architecture from
\cite{he:zhan:ren:sun:2015} which has 177 layers.
See Figure~\ref{fig:ResNetModel}.
We probed the network at
the outputs of the sixteen convolution / identity blocks,
corresponding to layers
$$\ell = 17, 27, 37, 49, 59, 69, 79, 91, 101, 111, 121, 131, 141, 153, 163, 173.$$
The accuracy attained by adjoining a softmax layer after layer $\ell$
is plotted versus $\ell$ in Figure \ref{fig:intermediateAccuracy}.
There we do not see the rapid initial gain in accuracy versus
$\ell$.  We attribute this difference to that network having been initialized
with values from training on ImageNet \cite{deng:dong:soch:li:li:fei:2009}.

\begin{table}
\centering
\begin{tabular}{r rrrrrrrrrr}
\toprule
& \multicolumn{10}{c}{Predicted class}\\
\midrule
True class & plane & car & bird & cat & deer & dog & frog & horse & ship & truck \\
\midrule
plane & 924 &   2 &  20 &   4 &   3 &   0 &   1 &   3 &  36 &   7 \\
car   &   3 & 972 &   0 &   0 &   0 &   1 &   0 &   0 &   5 &  19 \\
bird  &  13 &   0 & 929 &  11 &  16 &  10 &  11 &   5 &   5 &   0 \\
cat   &   5 &   0 &  24 & 848 &  12 &  88 &  14 &   3 &   4 &   2 \\
deer  &   3 &   0 &  12 &  16 & 946 &  10 &   5 &   8 &   0 &   0 \\
dog   &   2 &   2 &  11 &  68 &  11 & 895 &   2 &   6 &   1 &   2 \\
frog  &   4 &   1 &  14 &  12 &   2 &   5 & 961 &   0 &   1 &   0 \\
horse &   5 &   1 &   9 &   6 &   8 &  12 &   0 & 958 &   0 &   1 \\
ship  &   7 &   0 &   2 &   4 &   1 &   1 &   1 &   0 & 979 &   5 \\
truck &   9 &  26 &   0 &   2 &   1 &   2 &   0 &   1 &  12 & 947 \\
\bottomrule
\end{tabular}
\caption{\label{tab:confusioncifar}
This is the confusion matrix for the CIFAR10  test data.
}
\end{table}

\begin{table}
\centering
\begin{tabular}{rrrrrrrrrrrr}
\toprule
&\multicolumn{10}{c}{Predicted class}\\
\midrule
True class & t-shirt/top & trouser & pullover & dress & coat & sandal & shirt & sneaker & bag & ankle \\
\midrule
t-shirt/top & 873 &   1 &  12 &  12 &   4 &   1 &  94 &   0 &   3 &   0 \\
trouser     &   1 & 987 &   1 &   6 &   1 &   0 &   2 &   0 &   2 &   0 \\
pullover    &  15 &   1 & 879 &  10 &  43 &   0 &  51 &   0 &   1 &   0 \\
dress       &  14 &   6 &   7 & 919 &  26 &   0 &  26 &   1 &   0 &   1 \\
coat        &   0 &   0 &  55 &  24 & 878 &   0 &  43 &   0 &   0 &   0 \\
sandal      &   0 &   0 &   0 &   0 &   0 & 983 &   0 &  15 &   0 &   2 \\
shirt       &  86 &   3 &  47 &  19 &  69 &   0 & 768 &   0 &   8 &   0 \\
sneaker     &   0 &   0 &   0 &   0 &   0 &   8 &   0 & 973 &   0 &  19 \\
bag         &   3 &   2 &   1 &   1 &   3 &   1 &   2 &   3 & 983 &   1 \\
ankle boot  &   0 &   0 &   0 &   0 &   0 &   5 &   1 &  32 &   0 & 962 \\
\bottomrule
\end{tabular}
\caption{\label{tab:confusionfashion}
This is the confusion matrix for the FashionMNIST test data.
}
\end{table}

The confusion matrix for the CIFAR10 data is in Table~\ref{tab:confusioncifar}.
Separating cat images from dog images appears to be the hardest task.
Trucks and cars are also frequently confused with each other.  Many planes are labeled
as ships but the reverse is less common.
Table~\ref{tab:confusionfashion} shows the confusion matrix for the
FashionMNIST data.
Coats and pullovers are commonly confused with each other
as are ankle boots and sneakers, while shirts are confused with four other classes.

\section{Probing with t-SNE}\label{sec:tsne}

An alternative to training an $\ell+1$'st layer is to use t-SNE on the $N\times M_{\ell}$ matrix of output at layer $\ell$. The t-SNE algorithm produces a lower dimensional representation of the points, that like multidimensional scaling \cite{ torg:1952, krus:1978, cox:cox:2008} seeks to have low dimensional interpoint distances be representative of the original higher dimensional interpoint distances.
It also emphasizes smaller interpoint distances as in local multidimensional scaling \cite{chen:buja:2009}.
The algorithm proceeds without knowing the classifications of the data or even the number $K$ of levels of the categorical response.
Following \cite{derk:2016}
we projected the $M_\ell$ dimensional
points onto their first $50$ principal components before running t-SNE.
The t-SNE algorithm involves some randomness. Its criterion is also invariant
to rotations or reflections of the point clouds.
In order to make the images more nearly
comparable we used the same random seed at each layer.  Furthermore, the practice of
projecting the intermediate outputs down to the same dimension, 50, using PCA in each case makes the starting points for the layers
more closely aligned with each other than they would otherwise be.
The lone exception to the above description is for the output of layer 60 of the VGG15 model for the CIFAR10 dataset. This output is 10-dimensional as it is the output of the entire network with one value per class.  In this case, we use all 10 dimensions in the t-SNE algorithm.
Because t-SNE's criterion is dominated by small interpoint distances, the precise
location and orientation of a cluster well separated from the rest of the data
could be strongly influenced by the random seed.  The relative position of clusters
that overlap or nearly do is likely to be a more stable feature.

\begin{figure}
\begin{subfigure}{.95\textwidth}
\centering
\includegraphics[width=1.1\hsize]{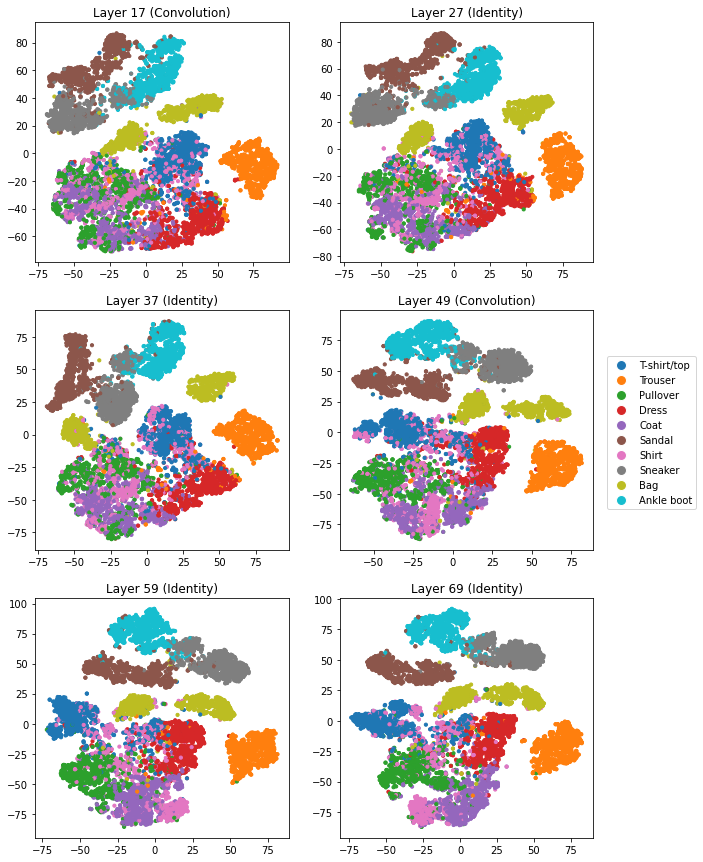}
\end{subfigure}
\caption{\label{fig:fashionmnisttsne1}
t-SNE output for FashionMNIST data after layers 17, 27, 37, 49, 59, and 69. The types of these blocks are listed above in the titles.
}
\end{figure}

\begin{figure}
\begin{subfigure}{.95\textwidth}
\centering
\includegraphics[width=1.1\hsize]{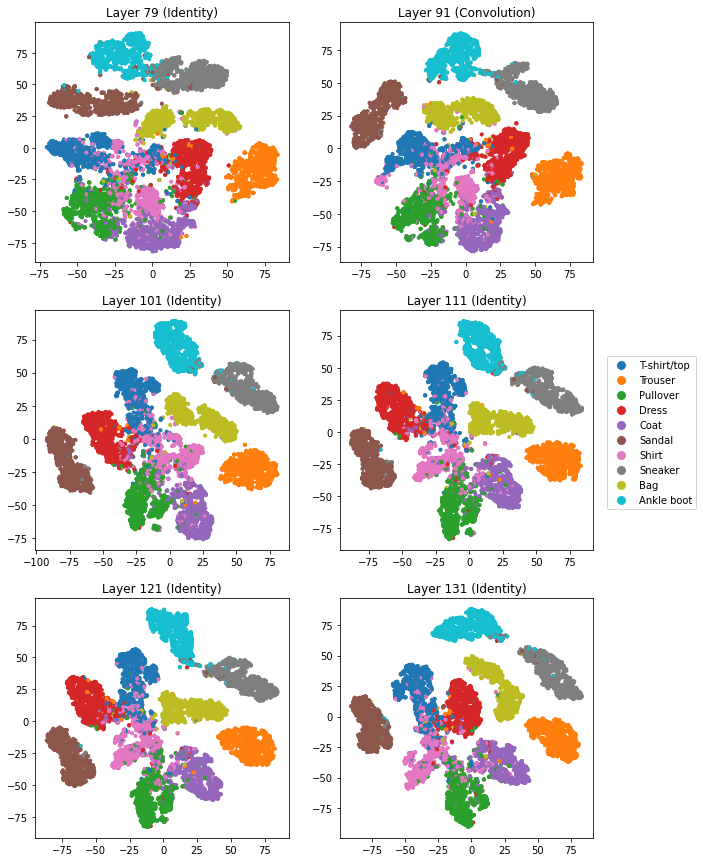}
\end{subfigure}
\caption{\label{fig:fashionmnisttsne2}
t-SNE output for FashionMNIST data after layers 79, 91, 101, 111, 121, and 131. The types of these blocks are listed above in the titles.
}
\end{figure}

\begin{figure}[t]
\begin{subfigure}{.95\textwidth}
\centering
\includegraphics[width=1.1\hsize]{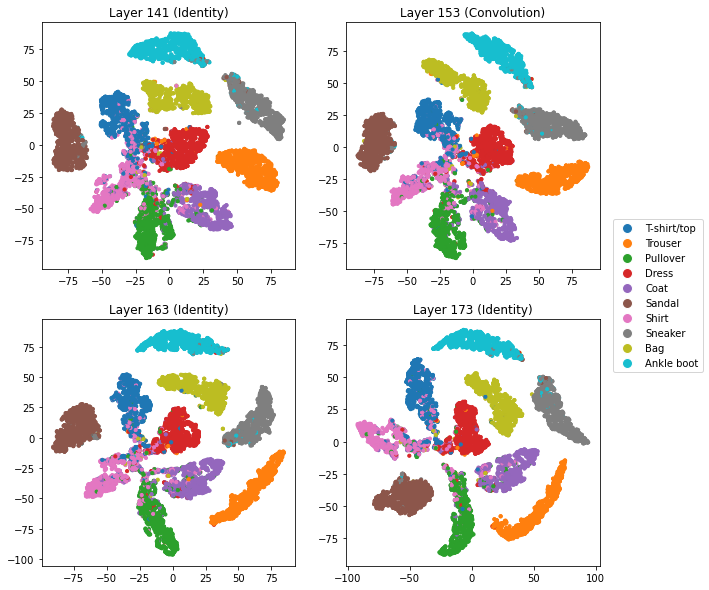}
\end{subfigure}
\caption{\label{fig:fashionmnisttsne3}
t-SNE output for FashionMNIST data after layers 141, 153, 163, and 173. The types of these blocks are listed above in the titles.
}
\end{figure}

Figures \ref{fig:fashionmnisttsne1}, \ref{fig:fashionmnisttsne2}, and \ref{fig:fashionmnisttsne3} show
t-SNE outputs for the fashion-MNIST data. These images depict the
held out, testing data, not the training data.
Up to and including layer 49, the images for `bag' clearly
plot as two disjoint and even widely separated clusters.
Inspection in Section~\ref{sec:ingroup} reveals that there are
two kinds of bag image, one that we could call purses and
another that mixes clutchs and satchels.
By purse images we mean those with handles/straps that are well separated
from and above the rest of the bag.  The clutchs don't have
straps, while the satchels have straps that
are nearly adjacent to the rest of the bag. Interestingly, by about layer 91 the two `bag' groups have
connected and by around layer 163 or 173 they appear to have merged into one.
The loss function used in training does not reward the network for noticing
that there are two kinds of bag and, as it happens, that distinction weakens or even disappears
from the t-SNE plot.
%If one is interested in looking for subgroups within classes
%then it may be best to inspect the results from intermediate layers.
%\textcolor{red}{I took this out. A reviewer might complain that
%the two sub-classes could be found by just clustering the original data.
% The stronger point is that we have evidence of the network forgetting something
% learned in earlier layers because it was not rewarded for knowing it.
%}

In Section~\ref{sec:ingroup}
we use a method of ranking images from most like their own class to least like their
own class.  We find that there are also two very different kinds of sandal images and possibly
two very different kinds of trouser images.
Figures~\ref{fig:histAgg_bag}, \ref{fig:histAgg_sandal} and \ref{fig:histAgg_trouser} there
depict examples of extreme bags, sandals and trousers.

We also see  in
Figures \ref{fig:fashionmnisttsne1}, \ref{fig:fashionmnisttsne2}, and \ref{fig:fashionmnisttsne3}
that the trouser and to a lesser extent bag images are already well separated
from the other classes by layer 17.
The footwear groups (sandals, sneakers, and ankle boots) start very connected with one another until
sandals separate by layer 69, and the sneakers and ankle boots separate from one another by layer 101.
Similarly, the upper body wear classes (t-shirt/tops, pullovers, dresses, coats, shirts) start out well
merged, and only really separate out from one another towards the very end at layer 173.

\begin{figure}
\begin{subfigure}{0.9\textwidth}
\centering
\includegraphics[width=1.1\hsize]{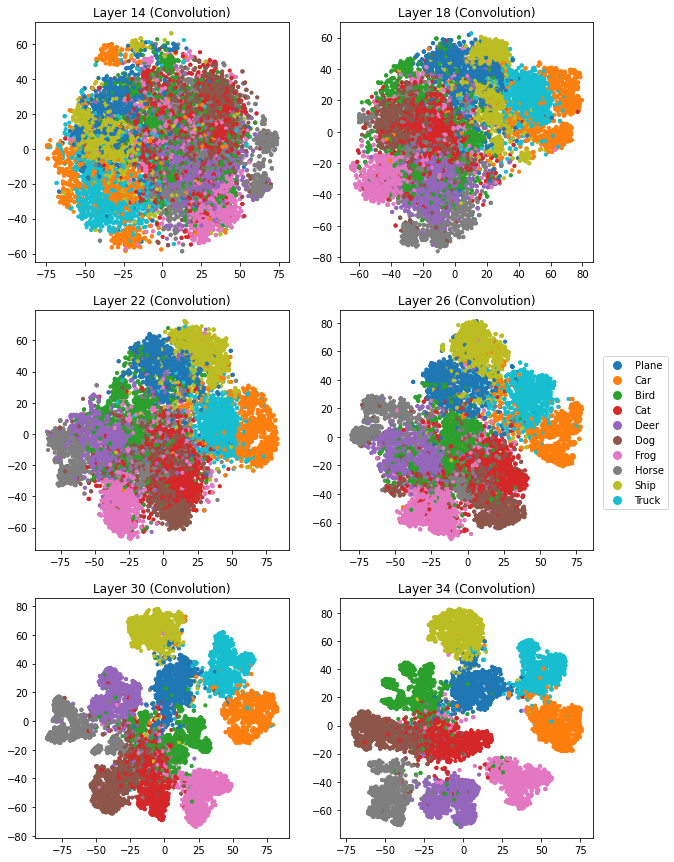}
\end{subfigure}
\caption{\label{fig:cifartsne1}
t-SNE output for CIFAR10 data after layers 14, 18, 22, 26, 30, and 34. The types of these layers are listed above in the titles.
}
\end{figure}

\begin{figure}
\begin{subfigure}{0.9\textwidth}
\centering
\includegraphics[width=1.1\hsize]{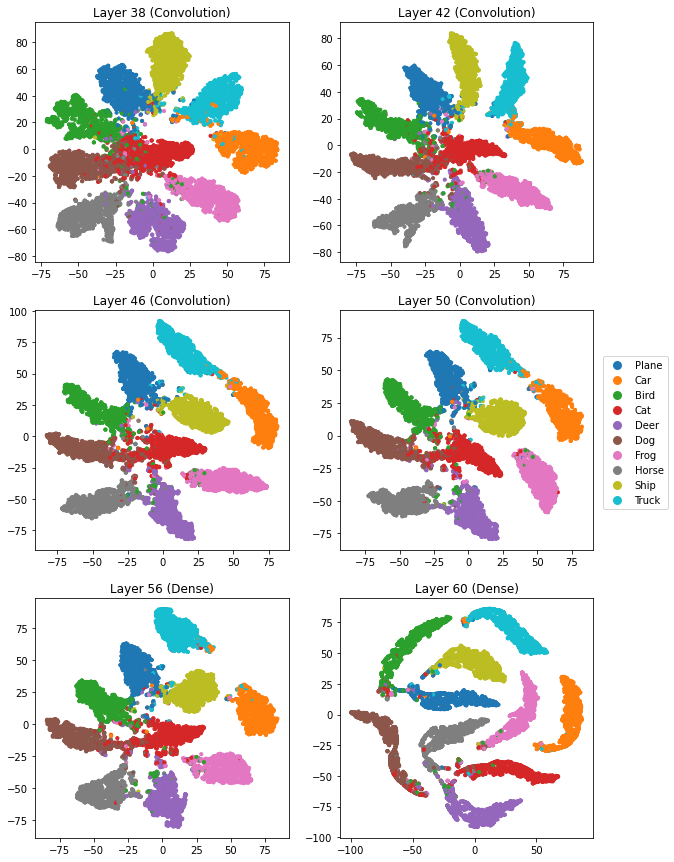}
\end{subfigure}
\caption{\label{fig:cifartsne2}
t-SNE output for CIFAR10 data after layers 38, 42, 46, 50, 56, and 60. The types of these layers are listed above in the titles.
}
\end{figure}

Figures~\ref{fig:cifartsne1} and~\ref{fig:cifartsne2} show the corresponding t-SNE plots
for the CIFAR-10 data at 12 different output layers ranging from layer 14 to layer 60.
A striking difference in these figures is that in the early layers (14 and 18)  the points are
not at all separated into disjoint clouds.  Recall that the fashion-MNIST network
was initialized to weights from training in ImageNet.  The network for CIFAR-10
by contrast was trained from a random start. By layer 30 we start to see groups
forming though there is still a lot of overlap.  Cats and dogs remain closely connected
up to and including layer 56, the second last one shown.
Recall that from the confusion matrices, separating cats from dogs is
especially hard.

By the final layer 60, the $10$ classes are very well
separated in a two dimensional view, forming
narrow and bent clusters.
There are still errors in the classification.
Many of those points plot within clusters for another
category, very often near one of the tips
and very often nearer the center of the image.
That latter pattern was more pronounced for FashionMNIST.
In the t-SNE for layer 60, cat and dog
clusters are well separated but some of the cat
images are at one end of the dog image cluster.
On inspection we find that these
points were misclassified but with high confidence.
%\textcolor{red}{ Art: This claim is true. While some of the points where the dog + cat clusters are adjoined are a bit confused between the two, as we move into the dog cluster, we mistake the images of cats as images of dogs with high confidence (at least 90\%).}
% Thanks for checking.

\section{Class specific vectors}\label{sec:classv}

While t-SNE is powerful and evocative, the spatial placement of
individual points is hard to interpret.  A principal components
analysis offers some alternative functionality.  Given the data,
the PC mapping is not random, so the position of points is more reproducible
and in particular, the closeness of clusters is not subject to algorithmic noise.
Second, having constructed principal components for one
set of data, we can project additional data onto them. For instance
we could define principal components using the training data
and, retaining those eigenvectors, project test data into the same space.

The data we use are $X\in\real^{N\times M}$ and $Y\in\{0,1,\dots,K-1\}^N$.
The matrix $X$ changes from layer to layer as does $M$,
while $Y$ remains constant.
Each column of $X$ represents the output from one neuron in
the layer under study.
We center these outputs by subtracting the
mean from each column of $X$.
That way we ensure that
$\sum_{i=1}^N\bsx_i=\bszero\in\real^M$
where $\bsx_i\in\real^M$ is the $i$'th row of $X$ expressed
as a column vector.

\begin{figure}
\begin{subfigure}{0.9\textwidth}
\centering
\includegraphics[width=1.1\hsize]{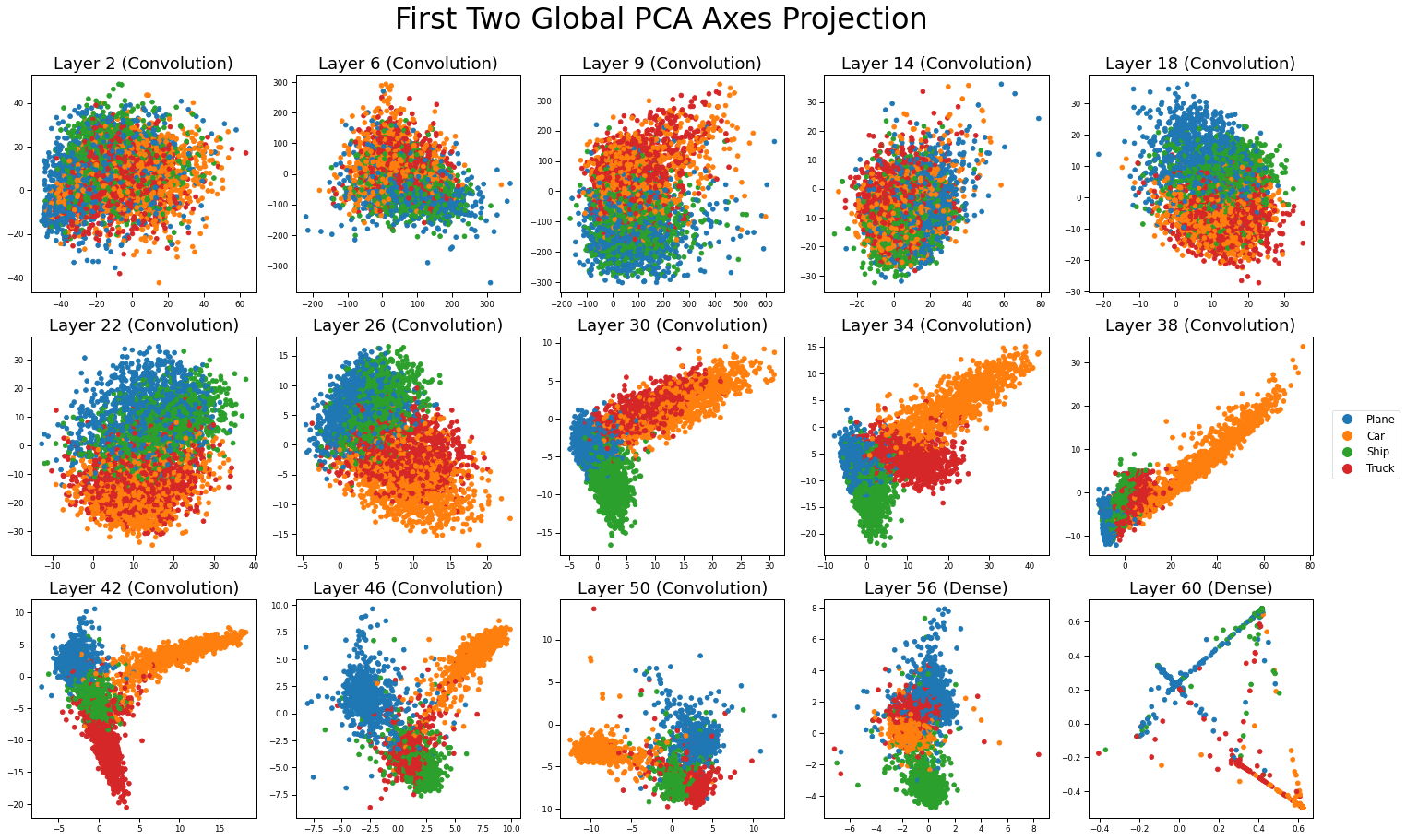}
\end{subfigure}
\caption{\label{fig:pairplots_globalPCA_aggregate}
Projections from the top two principal components in a global PCA for all activation layers, visualizing the planes, cars, ships, and trucks in CIFAR10.
}
\end{figure}

\begin{figure}
\begin{subfigure}{0.9\textwidth}
\centering
\includegraphics[width=1.1\hsize]{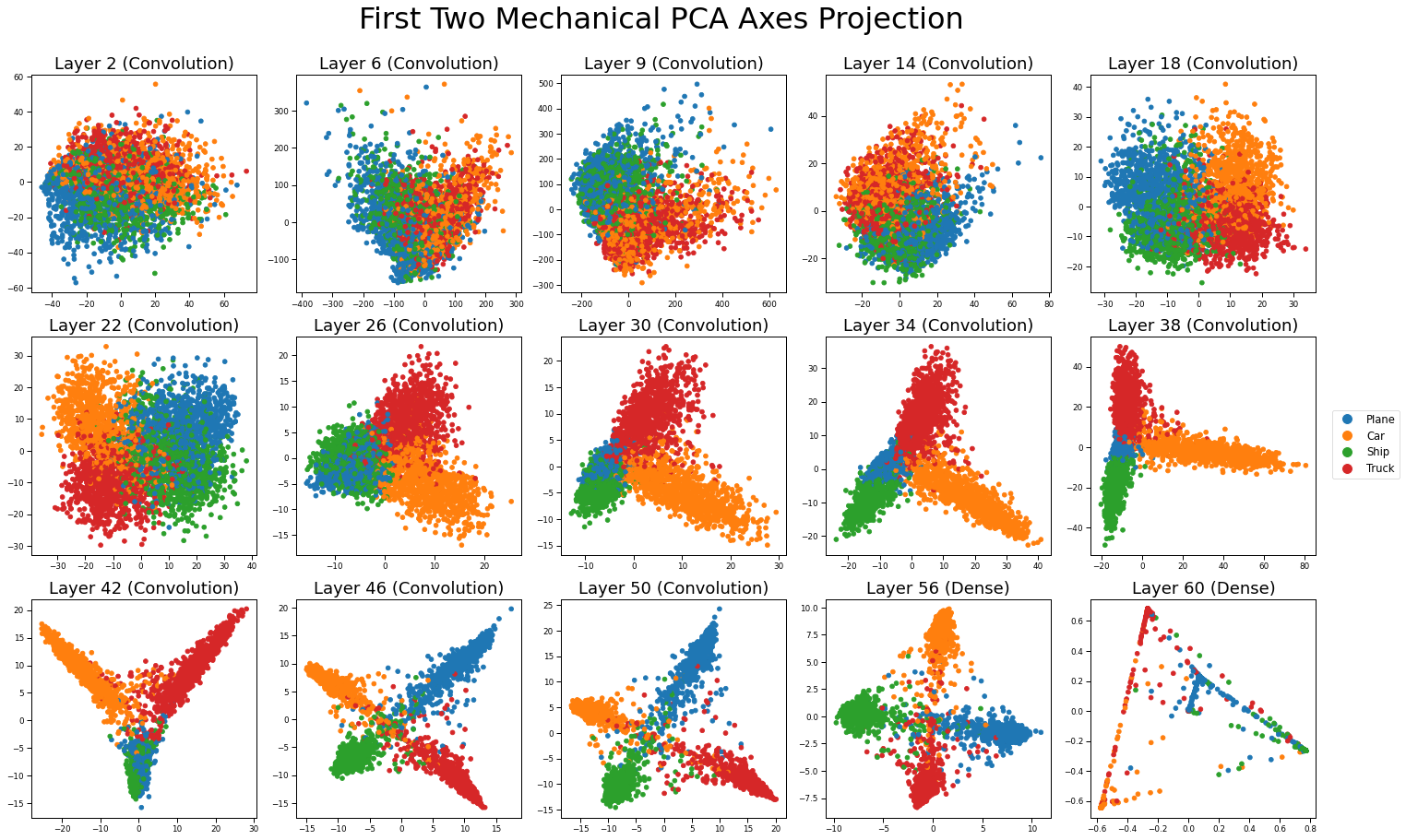}
\end{subfigure}
\caption{\label{fig:pairplots_mechPCA_aggregate}
Projections from the top two principal components in a PCA involving only the mechanical groups (planes, cars, ships, and trucks) for the outputs of all activation layers, visualizing the planes, cars, ships, and trucks in CIFAR10.
}
\end{figure}

We are interested to see the arrangements among commonly
confused classes. For instance, it is of interest to see
the relative arrangements among the mechanical images
because there are frequent errors within those classes.
Figure \ref{fig:pairplots_globalPCA_aggregate}
shows the global principal components plots for the
outputs of only the groups of ships, planes, cars, and trucks in CIFAR10.
What we see there is that projecting all ten classes at once yields
a lot of overlap.  We get a better view from PC plots
based on just those classes as in
Figure \ref{fig:pairplots_mechPCA_aggregate}.
There we see that in the later layers the four classes plot as
lengthy ellipsoids from a common origin.

We see that the last pre-softmax layers are
mostly consistent with neural collapse \cite{papy:han:dono:2020}
in that the vectors for most pairs of groups are nearly orthogonal
to each other by the later layers.  The clouds do extend towards
the origin where we see some overlap.

%\textcolor{red}{Art: the vectors start nearly parallel at the very start, and quickly move toward becoming orthogonal by halfway. The vectors usually have a cosine of about $\pm 0.2$ or so, but certain groups can pop larger. Pairs of groups with negative deviations tend to be groups that are similar in nature (dogs and cats are one example. Birds and planes are another.) Groups with positive deviations tend to be groups that are very dissimilar in nature (like cars and frogs). However, I'm not sure how much we want to cover this since I think we risk over-attributing significance to the values of the cosines.}

%\textcolor{red}{However, we also note that while for certain groups, like cars and trucks, the vectors are roughly orthogonal, we notice that the comets tend to bend into one another towards their bases. A similar effect arises for planes versus ships.}

%\textcolor{blue}{Chris: please check whether they're actually opposite
%and not just a tetrahedron that looks like opposites when projected
%into the plane. That might be in the cosines, but I'm not 100\%
%sure what vectors define the cosines.} %\textcolor{red}{Art: From the projections I can tell, it's definitely not a tetrahedron, but not exactly opposites always. The image I have is that of a flattened+warped simpled. Certain axes are opposite (see cars+trucks below), but most tend to be just orthogonal. The cosines point to that the axes are mostly orthogonal (the dot product is usually $\pm0.2$), but this doesn't always capture the behavior at the center of the comets.}
We also see that in layers 30 to 42, the data for images of planes
projects to nearly the origin in the first two principal components.
Only later do the planes clearly stand out from the other mechanical classes.
 Layer 60 reflects data after the softmax transformation and
it shows clearly that most points have at most one or two
non-negligible possibilities because the points are mostly
projected from a tetrahedral wireframe pattern.

It is a nuisance to redefine the principal components for every
subset of variables that we might wish to explore.
We prefer to define class-specific projections of the data.
For each $k\in\{0,1,\dots,K-1\}$ we choose a one-dimensional
projection especially representative of class $k$, as described
below,  and then view the data
in projections given by two of those vectors.
Class-specific vectors were studied by \cite{krza:1979}.
That work did not look at them for visualization.

Let $N_k$ be the number of indices $i$ with $Y_i=k$.
The matrix $\xk\in\real^{N_k\times M}$ has all the $\bsx_i$
for which $Y_i=1$.
% we may have to change notation later if/when X needs subscripts
Now we choose a special vector $\theta_k\in\real^M$
for cluster $k$.  There are several choices.
We could take $\theta_k$ to be the unit
vector proportional to $ (1/N_k)\sum_{i:Y_i=k}\bsx_i$
the mean vector for cluster $k$.
A second choice is the unit vector
$\theta_k=\arg\max_{\Vert\theta \Vert=1}\theta^\tran \xk^\tran\xk\theta$.
The third choice is the first principal component
vector of $\xk$.  This differs from the second
choice in that we must recenter $\xk$ to make
its rows have mean zero, yielding $\tilde\xk$.
Then $\theta_k$ maximizes
$\theta^\tran \tilde\xk^\tran\tilde\xk\theta$
over unit vectors $\theta\in\real^M$.
Below we work with the third choice.
The vectors we get characterize the principal direction of points
within each point cloud, not differences in the per class means.
%\textcolor{red}{Correct}

When we take the within-class principal component vector
we have also to choose its sign. If $\theta$ is the first
principal component vector then $-\theta$ has an
equal claim to that label. We choose the sign so that
the mean of the data for class $k$
has a larger inner-product with $\theta$ than
%the corresponding inner-product between
the global mean has. %and $\theta$.

\begin{figure}
\begin{subfigure}{0.9\textwidth}
\centering
\includegraphics[width=1.1\hsize]{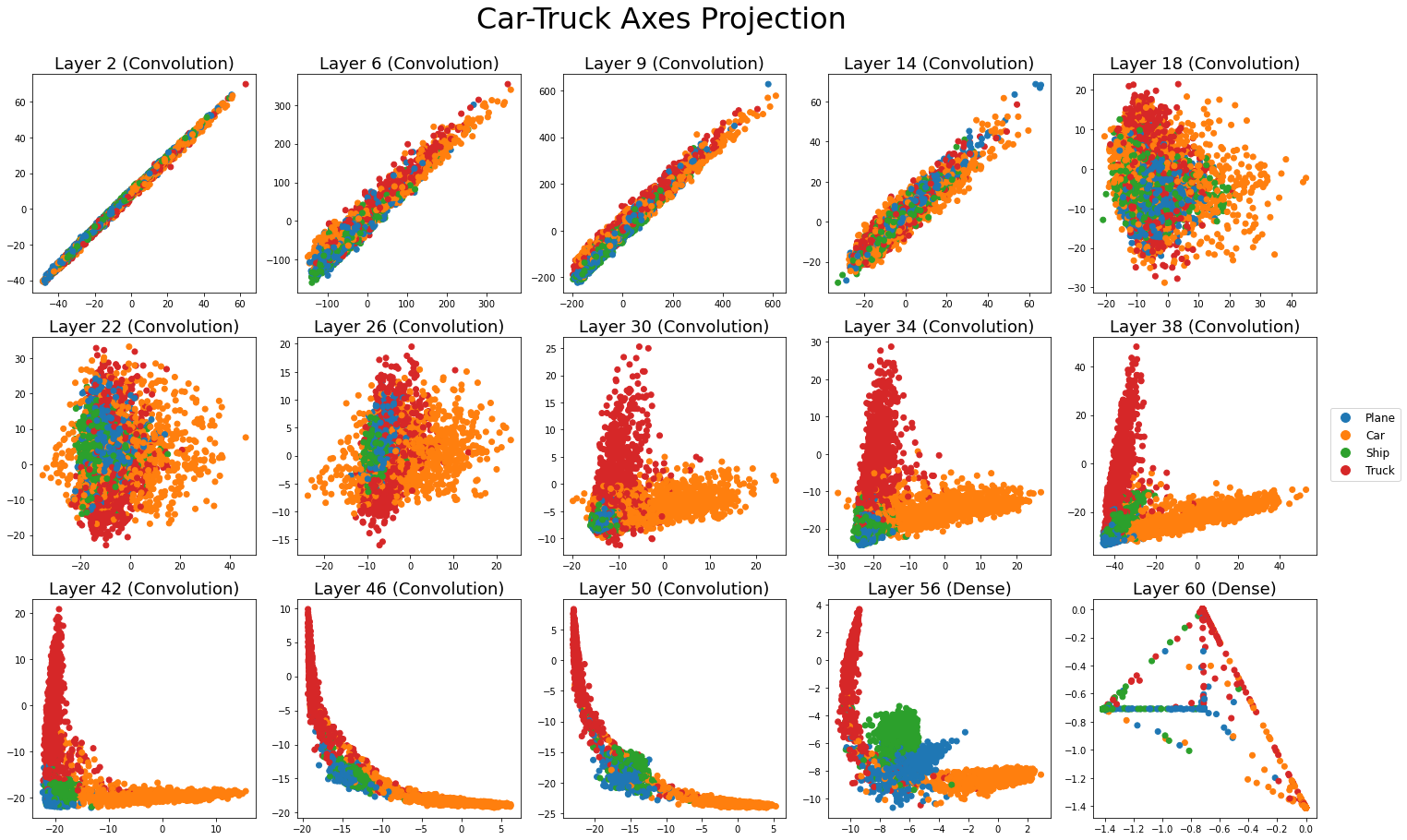}
\end{subfigure}
\caption{\label{fig:pairplots_categoryPCA_aggregate}
In each plot, the vertical axis has data projected on the class-specific vector for `car' images
and the horizontal axis uses the class-specific vector for `truck' images.
The points are for the plane, car, ship, and truck images in CIFAR10.
%\textcolor{red}{We note that the principal axes for cars and trucks start off approximately parallel at the very start, and switches to become roughly orthogonal at layer 18. We also observe that in layers 46, 50, and 56, while the car and truck axes are nearly perpendicular, the clusters of points for these two groups "bend" into one another where they meet. }  % moving this remark to body of paper and editing it a bit
}
\end{figure}

Using class-specific projections, we make
scatterplots of $X\theta_k\in\real^N$ versus $X\theta_{k'}$
for pairs of classes $k\ne k'$.
These plots show one point per sample.
Figure~\ref{fig:pairplots_categoryPCA_aggregate}
shows the mechanical images plotted in the
space defined by the car versus plane vectors.
This is a projection but not an orthogonal projection
into $\real^2$ because $\theta_k$ and $\theta_{k'}$
need not be orthogonal.
In the early layers, the class-specific
projections produce highly correlated scatter plots.
This is largely because the vectors
$\theta_k$ and $\theta_{k'}$ start out correlated.
In later layers, points vary greatly in the projections from their own class
and much less in the other projections.
The class-specific asxes for cars and trucks start off approximately parallel
start to become nearly orthogonal by layer 18 (though the point clouds
still overlap a lot).
By layer 30 we see meaningful separations among the points.
In layers 46 and 50 it appears that the angle between those
clouds is more than 90 degrees.  This is confirmed by
a  tour image (in Section~\ref{sec:classt})
that makes an orthogonal projection
of the data into the plane spanned by those two vectors.
We also observe that in layers 46, 50, and 56, while the car and truck axes are nearly perpendicular, the clusters of points for these two groups ``bend" into one another where they meet.   % moving this remark to body of paper and editing it a bit

Car and truck images are fairly similar.
At layer 30, we see that trucks tend to plot at positive
values along the car axis.
By layer 46, trucks are now plotting at negative values
along the car axis.  The intervening layers have
been configured to push those classes farther apart.
Planes and ships do not proceed through such a process.
Instead they have near zero projections along both
car and truck directions through training.

%\textcolor{blue}{Here is where it is important to make sure that we are not fooling ourselves via our choice of $\pm\theta$.}

\section{Class specific guided tour}\label{sec:classt}

A tour of $X\in\real^{N\times M}$ is an animated sequence of projections (also commonly referred to as frames) of $X$ into $\real^{N\times d}$, where $d$ is typically a small value such as 2 or 3.   The best known of these is the grand tour \cite{asim:1985}
which generates a sequence of such projections designed to
explore the space of all projections by eventually getting close to every point in the manifold of projections from $\real^M$ to $\real^d$.
That manifold is quite large, so it can take a long time to get close
to every view.  Therefore specific tours have been developed to make a more focussed exploration
of views.
Simple tours generally consist of moving along a sequence of projections using
interpolated
paths between sequential pairs of projections of interest. The resulting animation rotates smoothly between these pairs. The technical procedure is outlined in
Section 4.1 of Buja et al. \cite{buja:cook:asim:hurl:2005}.
% this paper is not using author-year citations
Planned tours refer to tours where the (usually finite) sequence of frames are predetermined.

%\sout{ To visualize the data, we choose a guided tour that cycles through a preset list of frames defined by the spans of pairs of the $\theta_k$ tours that explore the space . The frame defined by $\theta_k$ and $\theta_{k'}$ is an orthogonal projection onto the space that they span. As noted above those two vectors are not orthogonal in this projection. We find that these class-specific frames create visualizations that are much more meaningful than choosing frames that are chosen randomly. }

To visualize the data, we choose to explore the space created by the span of the ten class-specific vectors. For ease of computation, we fit a PCA with a single component to the class specific data $X_k \in \mathbb R^{N_k \times M}$ to acquire the principle axes $\theta_k \in \mathbb R^M$ and the decomposition of the data onto the class-specific axis $(X - \bsone\bar x_k^\tran) \theta_k$, where $\bsone\in\real^N$ is the all ones vector.. At this point, we flip the value of $\theta_k$ via products with negative one so that the class mean has a larger dot product with $\theta_k$ than the global mean, or equivalently that $(\bar x - \bar x_k) ^\tran\theta_k < 0$ where $\bar x \in \mathbb R^M$ is the global mean. Aggregating this information, we then have collected $X' = [ (X-\bsone\bar x_1^\tran) \theta_1, \ldots, (X-\bsone\bar x_K^\tran) \theta_K ] \in \mathbb R^{N \times K} $ as well as $\Theta = [\theta_1, \ldots, \theta_K]$. We then center $X'$ so that the columns have means zero, producing $\tilde X' = (X - \bsone\bar x^\tran) \Theta$, which is used as the data to produce the class-specific pair plots.

% leverage was in the same sentence twice
To produce the tours that leverage the class specific vectors $\theta_k$, we apply the technique used in
\cite{buja:cook:asim:hurl:2005} and project the data $X$ into the space spanned by the $K$ class-specific vectors since $K$ is usually much smaller than $M$. Note that we cannot use tours on $(X - \bsone\bar x^\tran) \Theta$ directly since this is not an orthogonal projection of the original centered data $X - \bsone\bar x^\tran$. To get around this, we perform the QR-decomposition $\Theta = Q_\theta R_\theta$ with $Q_\theta \in \mathbb R^{M \times K}$ and $R_\theta \in \mathbb R^{K \times K}$. We then visualize the projection of the data $ (X - \bsone\bar x^\tran) Q_\theta $. If $K$ is much smaller than $M$ and that the class-specific vectors are linearly independent, we can compute this efficiently as $(X - \bsone\bar x^\tran) \Theta \cdot R_\theta^{-1}$ since we already have $(X - \bsone\bar x^\tran) \Theta$ from the computation for the class specific pair plots.

%(The above information might go better above when discussing the pair plots...)

In the case that the $K$ class-specific vectors are not linearly independent, special considerations need to be made. While most layers do not have this issue,
the output layers, such as layer 60 in the VGG15 model, ordinarily do.
This is because the output is 10-dimensional with the constraint that the ten components need to add up to one, and thus the space spanned by the ten class specific axes is at most 9-dimensional. In this case, we again compute the QR-decomposition $\Theta = Q_\theta R_\theta$, but we now force $Q_\theta \in \mathbb R^{M \times r}$ and $R_\theta \in \mathbb R^{r \times K}$ where $r$ is the rank of $\Theta$ and calculate $(X - \bsone\bar x^\tran) Q_\theta$ directly.

In either case, we note that $R_\theta = Q_\theta^\tran \Theta$ implies that the columns of $R_\theta = [r_1, \ldots, r_K]$ are the projections of each of the class-specific vectors to the space spanned by all of the class-specific vectors. Doing so, we can condense all information about tours in the space spanned by the class specific vectors down to the projections $(X - \bsone\bar x^\tran) Q_\theta$ and $R_\theta = Q_\theta^\tran \Theta$ of the data and the class specific vectors respectively.

\begin{figure}
\begin{subfigure}{0.9\textwidth}
\centering
\includegraphics[width=0.8\hsize]{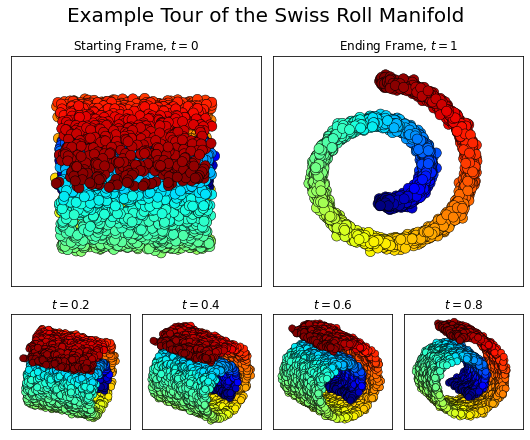}

\end{subfigure}
\caption{\label{fig:tourExamples} The plots show a possible path that could be seen on a tour on the Swiss Roll Manifold \cite{tene:silv:lang:2000}. We start from a top-down perspective at $t=0$ to a side-on perspective by $t=1$. At each $t \in [0,1]$ we move between these two perspectives with constant speed. }
\end{figure}

\begin{figure}[t]
\begin{subfigure}{0.9\textwidth}
\centering
\includegraphics[width=0.45\hsize]{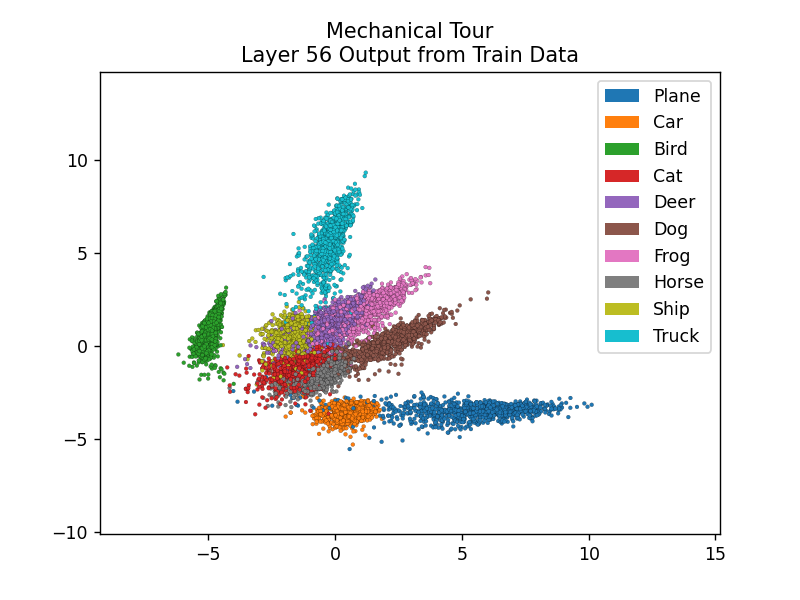}
\includegraphics[width=0.45\hsize]{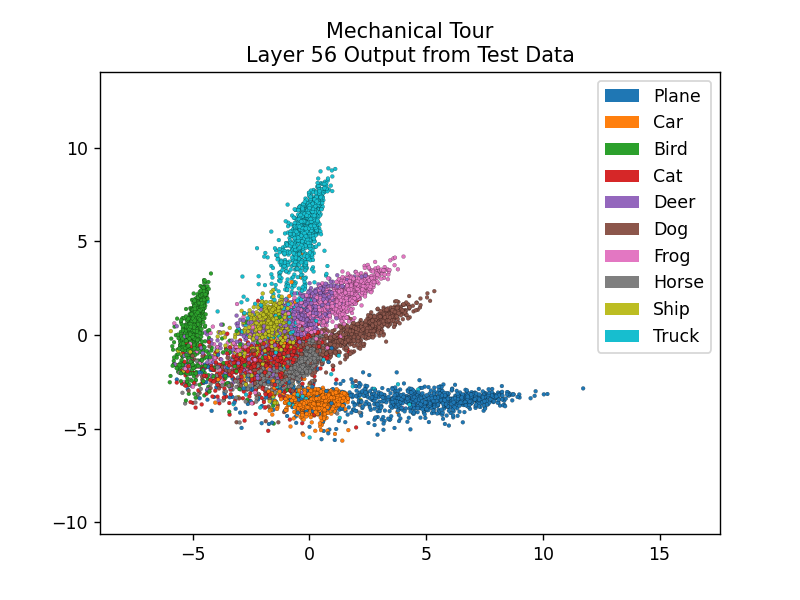}
\includegraphics[width=0.45\hsize]{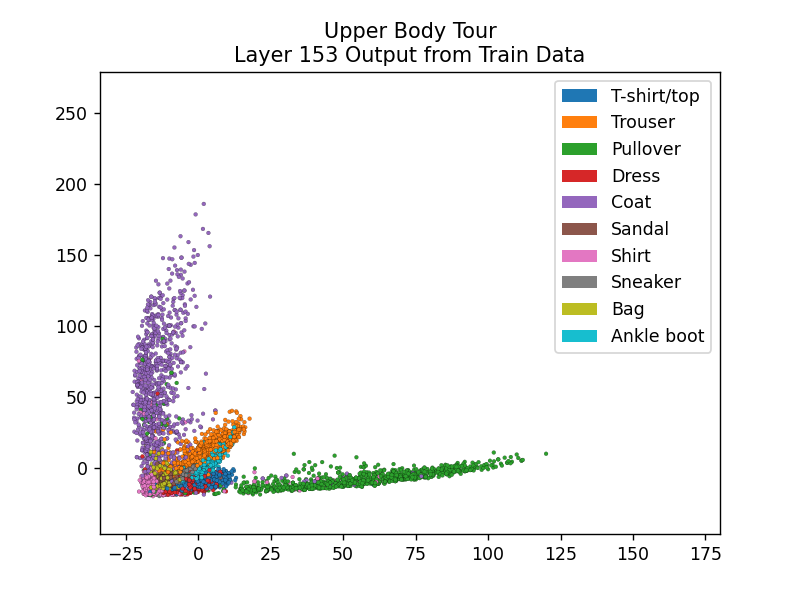}
\includegraphics[width=0.45\hsize]{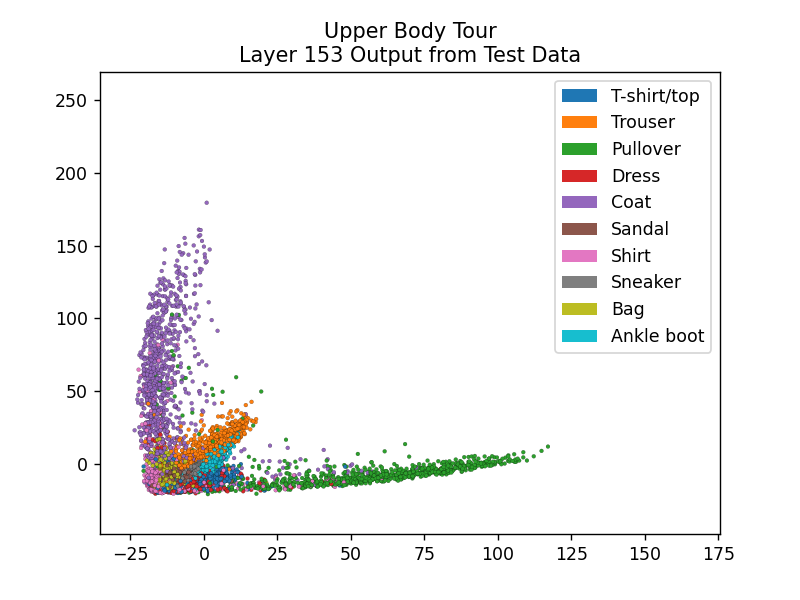}
\end{subfigure}
\caption{\label{fig:tourFrames} The plots show frames shown within the mechanical tour of the intermediate outputs of the VGG15 model (top row) and the ResNet50 model (bottom row). We can see several differences both between the two models as well as with the intermediate outputs of the training (left column) and testing data (right column).}
\end{figure}

For an animated view of the see
\begin{center}
\url{https://purl.stanford.edu/zp755hs7798}.
\end{center}
Figure \ref{fig:tourExamples} shows still images of frames that can be seen on an example toy Swiss Roll dataset.
Figure \ref{fig:tourFrames} shows frames that can be seen when exploring the VGG15 and ResNet50 models for both training and testing intermediate outputs.

%fig:tourFrames

% https://science.sciencemag.org/content/290/5500/2319.abstract?casa_token=8kwyh21vC9QAAAAA:TT5wKfLtfLg3OWxU_0pMJ-qYYovAm2WQraWwxlygnUgLcwRvnjgfyDQQd2fV0PSI89Xfh_1TjoL-

%  \textcolor{blue}{I think these bullets describe specific mp4s.  There are lots of them at the link. Please connect the bullet points to a way for somebody to find the specific mp4 and whatwe learn from seeing it. }

%\newpage

{%\color{red}
For the CIFAR10 dataset, we visualize planned two-dimensional tours inspired by the class specific vectors. One option is to consider a grand tour-like approach, where we generate a collection of two-dimensional frames by picking them uniformly at random from the set of such frames embedded in the space spanned by the ten class specific vectors. A second option is to restrict our attention to two-dimensional frames that are generated from pairs of the class-specific vectors by %\textcolor{red}{\sout{performing a QR-decomposition on $[\theta_j, \theta_k]$, where $\theta_j$ and $\theta_k$ are the class specific vectors corresponding to the pair of interest}
using frames of the form $[\theta_j, \theta_k']$ where $\theta_k'$ is the component of $\theta_k$ that is orthogonal to $\theta_j$ in order to explore the centered data $X - \bar{x}$. Note that we can equivalently use frames of the form $[r_j, r_k']$ where $r_k'$ is the component of $r_k$ orthogonal to $r_j$ to explore $(X - \bsone\bar{x}^\tran) Q_\theta$ for better computational efficiency.} For example, we consider a mechanical tour, where we use the frames generated by the pairs plane versus truck, car versus truck, car versus ship, and plane versus ship.

In general, we see that the planned tour strategy works much better compared to the grand tour-like strategy in the sense that the planned tours creates animations that are more interpretable and show more interesting features. For example, for \href{https://stacks.stanford.edu/file/druid:zp755hs7798/fig_CIFAR10_GrandTour_layer56_test.mp4}{layer 56 in the grand tour-like approach}, we can see some key features of the network. We see that the ten groups form comet like formations that point away from a shared common core. However beyond this, it is difficult to tell any relational information between the clusters. In contrast, for \href{https://stacks.stanford.edu/file/druid:zp755hs7798/fig_CIFAR10_MechanicalTour_layer56_test.mp4}{layer 56 in the mechanical tour approach}, we see that the comets are much more clearly defined and we can extract additional information from the plots. For example, we notice that for each frame, all classes have comets that generally point upwards and to the right. We also note that there is quite a bit of curvature within particular pairs, like how the truck points appear to bend into the cars three seconds in. We also note that at this point, the other eight groups overlap quite a bit. In contrast, for the frame of trucks vs planes at the very start, we notice that the trucks and planes points form orthogonal comets, and the rest of the eight groups spread out much more from one another.

This can also be seen in layer 30 within the VGG15 model, where  \href{https://stacks.stanford.edu/file/druid:zp755hs7798/fig_CIFAR10_GrandTour_layer30_test.mp4}{the grand tour-like approach} has comet like formations moving away from a common core, but \href{https://stacks.stanford.edu/file/druid:zp755hs7798/fig_CIFAR10_MechanicalTour_layer30_test.mp4}{the planned tour} showcases how these spokes are typically orthogonal. For the car versus truck frame, we note that all of the eight other groups mainly cluster around the origin, meaning that they have negligible components in the class specific axes associated with cars and trucks. In comparison, for the plane versus ship axis, we see that most of the groups tend to stick to the origin. The lone exception is cars, which has a distinctly negative ship component. These are distinct features that are much harder to visualize within the grand tour-like approach.

We perform a similar visualization for FashionMNIST intermediate outputs within the ResNet50 model. Again we consider a grand tour-like approach, selecting randomly generated frames from the span of the ten principal axes. We also consider two planned tours where we look at frames generated by pairs of class specific vectors. The first tour is composed of pairs of classes that are rarely confused for one another: t-shirt/tops versus ankle boots, trousers versus ankle boots, trousers versus bags, and t-shirt/tops vs bags. The second tour is composed of classes that are associated with upper-body clothing: t-shirt/tops versus shirts, pullovers versus shirts, pullovers versus coats, and t-shirt/tops versus coats.

When we visualize these tours, we see that the same trends as the CIFAR10 data persist. For example, in \href{https://stacks.stanford.edu/file/druid:zp755hs7798/fig_FashionMNIST_GrandTour_layer141_test.mp4}{layer 141 of the grand tour-like approach}, we again notice that the different groups form comet like shapes. However, beyond this, it's hard to observe any trends about how the comets are oriented about one another in space. When we move towards the rarely confused and upper-body planned tours, we can see other interesting features within the data. For \href{https://stacks.stanford.edu/file/druid:zp755hs7798/fig_FashionMNIST_UpperBodyTour_layer141_test.mp4}{layer 141 of the upper-body tour}, we observe that bags and sandals have no component along the class specific vectors associated with t-shirt/tops, pullovers, shirts, or coats. However, we see that trousers, ankle boots, and sneakers all have fairly positive components along these vectors. Visualizing the \href{https://stacks.stanford.edu/file/druid:zp755hs7798/fig_FashionMNIST_RarelyConfusedTour_layer141_test.mp4}{rarely confused planed tour of layer 141}, we notice that for the frame generated by the ankle boot and trouser class specific vectors, coats have a very large component in both of these vectors, while most other groups have components much closer to zero. We also notice that when we shift to the frame generated by bags and trousers that unlike other groups, bags will form two comet like clusters pointed in opposite directions. These features seen in the planned tours are generally harder to discover within the grand tour-like approach.

There are major differences between the VGG15 and ResNet50 models. One major difference is in how the intermediate points for the testing data compare to the training data. In the VGG15 model, we notice that in later layers that the class specific clusters from training data will form comets pointing away from a shared core, but their tails will overlap with one another. In contrast, the training data will have a similar shape and position, but their tails will not reach as far down and thus do not overlap. This can be seen prominently in layer 50 of the mechanical tour amongst the \href{https://stacks.stanford.edu/file/druid:zp755hs7798/fig_CIFAR10_MechanicalTour_layer50_test.mp4}{testing data} and \href{https://stacks.stanford.edu/file/druid:zp755hs7798/fig_CIFAR10_MechanicalTour_layer50_train.mp4}{training data}. This same effect can be seen to a much more limited extent in the ResNet50 model for layer 153 for the \href{https://stacks.stanford.edu/file/druid:zp755hs7798/fig_FashionMNIST_UpperBodyTour_layer153_train.mp4}{training data} and the \href{https://stacks.stanford.edu/file/druid:zp755hs7798/fig_FashionMNIST_UpperBodyTour_layer153_test.mp4}{testing data} for the pullovers class, but the effect is not as common among the other groups and layers. This is likely an indicator that the VGG15 model is more overtrained compared to the ResNet50 model.

A second difference is in the shape of the arms. We notice that in general, the comet-shapes for the class specific clusters for the ResNet50 model are fairly sharp and extend out fairly far. Specifically, we see fairly sharp arms for \href{https://stacks.stanford.edu/file/druid:zp755hs7798/fig_FashionMNIST_RarelyConfusedTour_layer141_train.mp4}{the rarely confused tour on the intermediate training output for layer 151} and for \href{https://stacks.stanford.edu/file/druid:zp755hs7798/fig_FashionMNIST_RarelyConfusedTour_layer163_train.mp4}{layer 163.} We note that the arms frequently extend out as more than 100 away from the origin, and become extremely sharp, especially on layer 163. In contrast, with \href{https://stacks.stanford.edu/file/druid:zp755hs7798/fig_CIFAR10_MechanicalTour_layer42_train.mp4}{layer 42} and \href{https://stacks.stanford.edu/file/druid:zp755hs7798/fig_CIFAR10_MechanicalTour_layer56_train.mp4}{layer 56} of the mechanical tour on the intermediate training data, we notice that the arms only extend out to at most
40, and the arms are much thicker. This is especially pronounced in layer 56 where the comets are much shorter and wider.

The final difference is in how fast the arms form. We notice that in the ResNet50 model, the arms have formed as early as \href{https://stacks.stanford.edu/file/druid:zp755hs7798/fig_FashionMNIST_RarelyConfusedTour_layer049_train.mp4}{layer 49 for the training data in the rarely confused tour}, which is only about a fourth of the way through the model. Although some classes haven't well differentiated well from others, some groups like bags and trousers have already spread away and formed arms away from the center. In contrast, the arms develop much later in the VGG15 model. We only see the arms clearly develop by about \href{https://stacks.stanford.edu/file/druid:zp755hs7798/fig_CIFAR10_MechanicalTour_layer30_train.mp4}{layer 30 for the training data in the mechanical tour}, which is about halfway through the model.

% Sharp arms for the ResNet50 model later on, chubbier arms for the VGG15 model
% Resnet50 has arms that "curve into" one another, VGG15 has arms that meet at much sharper angles.

\section{In-group Clustering}\label{sec:ingroup}

In our t-SNE plots we saw strong evidence of two kind of bag images.
That evidence was present in intermediate layers but had disappeared
by the final layers.  In this section we use class-specific projections
to make a systematic search for such subclusters and also identify
examples of extreme images so a human can interpret them.

At each intermediate layer we project the observations from class $k$ on
the vector $\theta_k$ to rank images from most typical of their class
to least typical. We normalize $N_k$ values
to have mean zero  variance one when taken over all ten classes,
not just class $k$.
so that they have mean square one.

\begin{figure}[t!]
\begin{subfigure}{0.9\textwidth}
\centering
\includegraphics[width=1.1\hsize]{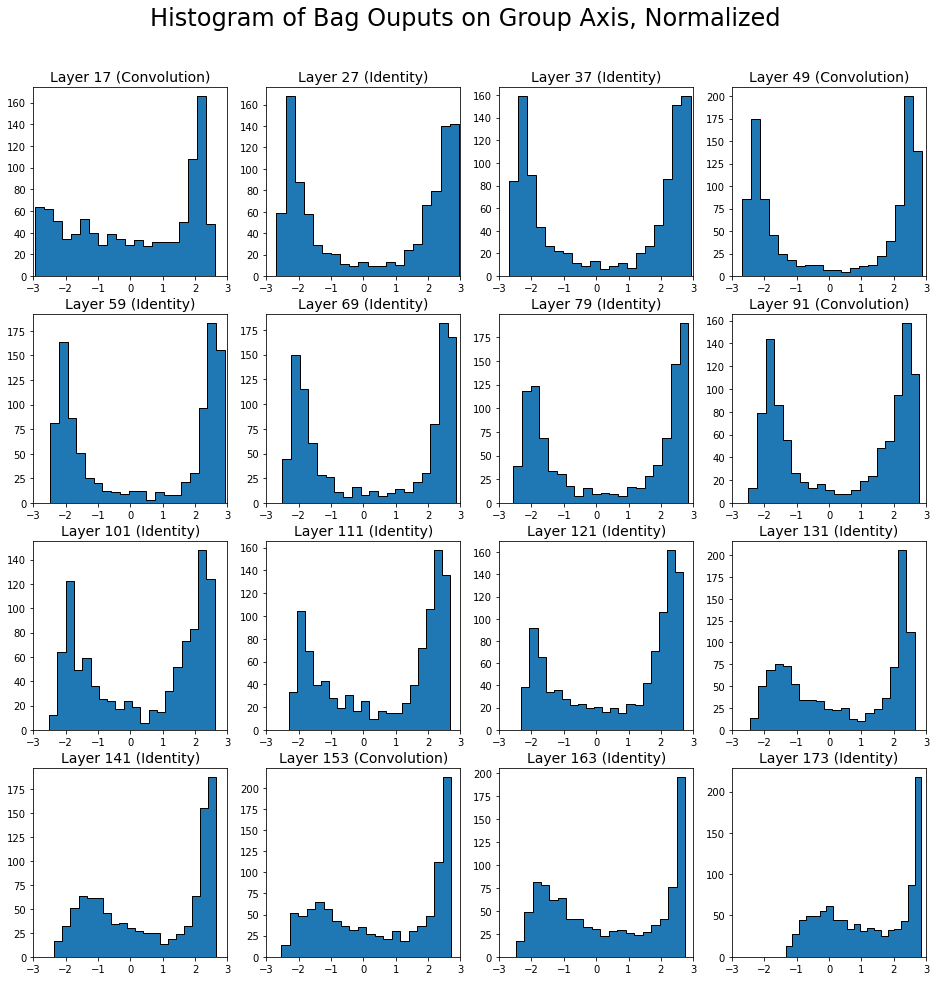}
\end{subfigure}
\caption{\label{fig:histAgg_bag}
Histograms of the intermediate outputs for test images labeled as bags for the outputs of the 16 residual units in the FashionMNIST network.
}
\end{figure}

\begin{figure}[t]
\begin{subfigure}{0.9\textwidth}
\centering
\includegraphics[width=1.1\hsize]{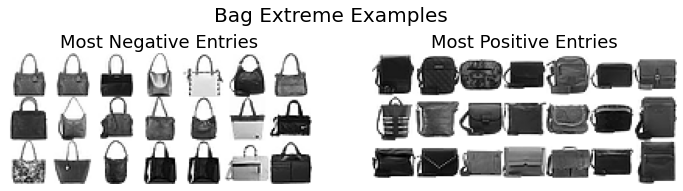}
\vspace{0.5cm}
\end{subfigure}
\begin{subfigure}{0.9\textwidth}
\centering
\includegraphics[width=1.1\hsize]{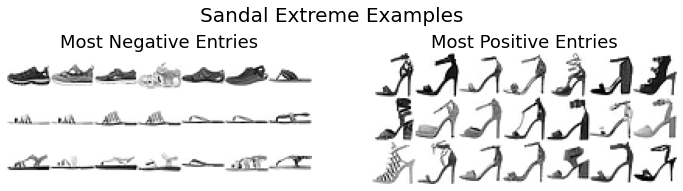}
\vspace{0.5cm}
\end{subfigure}

\begin{subfigure}{0.9\textwidth}
\centering
\includegraphics[width=1.1\hsize]{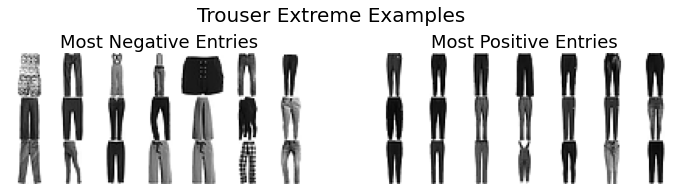}
\end{subfigure}
\caption{\label{fig:extremeExamples}
The most extreme examples of three groups along their group specific axis as intermediate outputs of the last residual unit (layer 173) for FashionMNIST.
}
\end{figure}

Figure~\ref{fig:histAgg_bag} shows those histograms for the images of bags.
A bimodal pattern becomes clear by layer 27 if not earlier, and
then becomes quite pronounced.
By final layer the modes are not very separated, nearly touching.
Recall that in the t-SNE plot those points had merged into a single cluster.
The top panel in Figure~\ref{fig:extremeExamples} shows images
of the bags at the extreme ends of the histogram for the final layer.
These represent two quite different kinds of image.  Although
the final histogram is less bimodal than the earlier ones we still
see two very different image types.
It is intuitively reasonable that these bag images look different
in early network layers that are forming image features.
It is also plausible that a network trained to distinguish
bags from other image types without regard to what
type of bag they are would in its later layers treat
them similarly.  It would take further research to understand
why the gap closes so late in the pipeline instead of earlier.

\begin{figure}[t]
\begin{subfigure}{0.9\textwidth}
\centering
\includegraphics[width=1.1\hsize]{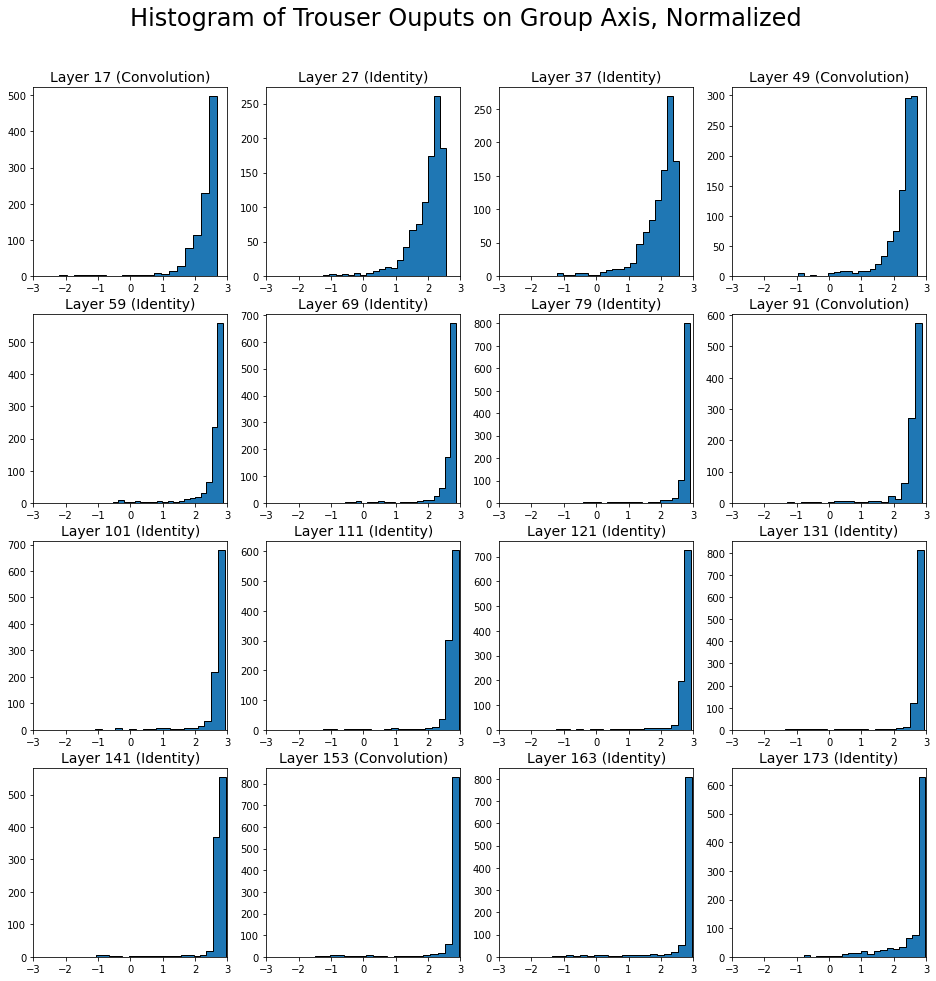}
\end{subfigure}
\caption{\label{fig:histAgg_trouser}
Histograms of the intermediate outputs for test images labeled as trousers for the outputs of the 16 residual units in the FashionMNIST network.
}
\end{figure}

The class-specific projection gives us a tool to look for subclasses.
Figure~\ref{fig:histAgg_trouser} shows histograms for trouser
images.  There we see no bimodal structure
and Figure~\ref{fig:extremeExamples} shows no especially striking
difference between the most and least typical trouser images.

\begin{figure}[t]
\begin{subfigure}{0.9\textwidth}
\centering
\includegraphics[width=1.1\hsize]{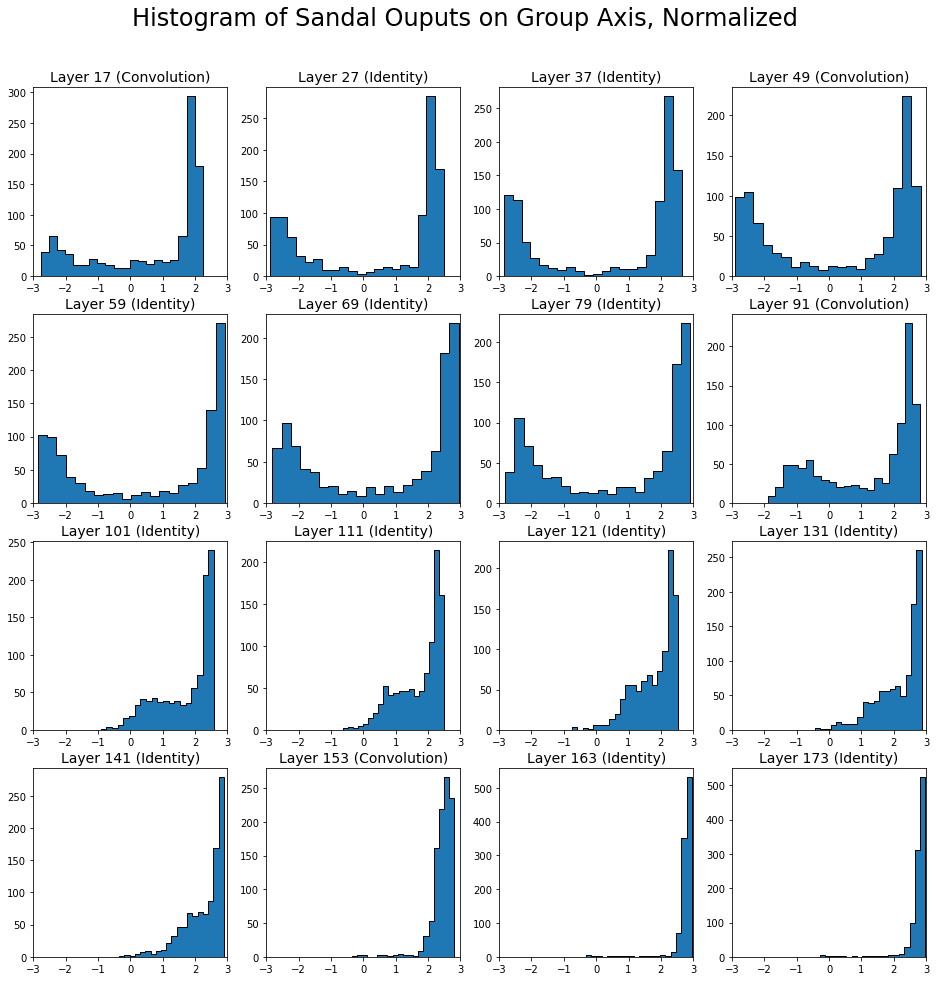}
\end{subfigure}
\caption{\label{fig:histAgg_sandal}
Histograms of the intermediate outputs for test images labeled as sandals for the outputs of the 16 residual units in the FashionMNIST network.
}
\end{figure}

Figure \ref{fig:histAgg_sandal} shows layer by layer histograms
for images of sandals.
The sandal images are bimodal at intermediate layers despite
the t-SNE images not placing the images into two clear groups.
As with bags, these histograms become much less bimodal
at later layers.  The bimodality disappears much earlier
for sandals than for bags.
The histograms for sandals become more skewed.
From Figure~\ref{fig:extremeExamples} we see
a sharp distinction between the types of sandals.
The histogram is distinguishing flat ones from high healed ones.

In this section we have looked for two kinds of image
by checking the extreme entries in a linear ordering.
We expect that more elaborate clustering methods
could work when an image class has three or more
types.  Our main point is that looking at intermediate
layers gives additional insights.

\section{Conclusions}

We have presented some graphical methods for probing neural networks layer by layer.
In doing so we saw that some clustering within a class can become muted or even
disappear as data progress through layers.

We devised class-specific projections
and found that they can help identify differences among image types within a class.
We had to make several choices in defining class-specific projections.
We believe that our choices are reasonable but we also
believe that other choices could reveal interesting phenomena.

The class-specific projections also provide very
focussed looks at point clouds when incorporated
into dynamic projections (tours).
We saw a big difference between the organization of early layer outputs when comparing
a network that was trained using transfer learning from ImageNet and one that was trained
from a random start.  We saw a variant of neural collapse wherein the outputs form ellipsoidal
clusters instead of spherical ones.
%o, and, some hard to resolve classes data are separated by closer to 180 degrees than 90.

\section*{Acknowledgments}

This work was supported by a grant from Hitachi Limited and by the US National Science Foundation under grant IIS-1837931. We thank Masayoshi Mase of Hitachi for comments and discussions about explainable AI.

\bibliographystyle{plain}
\bibliography{probing}

% I included but nothing came up
\iffalse
\include{pairplots_global}
\newpage
\include{pairplots_mech}
\newpage
\include{pairplots_category}
\fi

\end{document}